\documentclass{article}

\PassOptionsToPackage{numbers,compress}{natbib}
\usepackage[preprint]{neurips_2026}
 
\usepackage[utf8]{inputenc}
\usepackage[T1]{fontenc}
\usepackage{hyperref}
\usepackage{url}
\usepackage{relsize}
\usepackage{booktabs}
\usepackage{amsfonts}
\usepackage{amsmath,amssymb}
\usepackage{nicefrac}
\usepackage{microtype}
\usepackage{xcolor}
\usepackage{graphicx}
\usepackage{multirow}
\usepackage{xspace}
\usepackage{array}
\usepackage{rotating}
\usepackage{multirow}
\usepackage{makecell}
\usepackage{pifont}
\usepackage[capitalize,noabbrev]{cleveref}
\usepackage{tikz}
\usetikzlibrary{positioning,calc,fit,shapes.misc}
\usepackage{adjustbox}
\usepackage{ifthen}

\input{macros}
\newboolean{lowfigres}
\setboolean{lowfigres}{false}

\title{\textit{\modelname}: Bringing Hyperspectral Imagery into Multimodal Earth Observation Pretraining}

\author{%
  \textbf{Nassim Ait Ali Braham}$^{1,2}$,~
  \textbf{Aaron Banze}$^{2,3}$,~
  \textbf{Conrad M. Albrecht}$^{2,4}$, \\[0.3em]
  \textbf{Julien Mairal}$^{6}$,~
  \textbf{Jocelyn Chanussot}$^{6}$,~
  \textbf{Xiao Xiang Zhu}$^{1,5}$ \\[0.6em]
  \small
  $^1$Chair of Data Science in Earth Observation, Technical University of Munich, Germany \\
  \small
  $^2$Remote Sensing Technology Institute, German Aerospace Center (DLR), Germany \\
  \small
  $^3$Department of Aerospace Engineering, University of the Bundeswehr Munich, Germany \\
  \small
  $^4$LEAP, Columbia University, USA \\
  \small
  $^5$Munich Center for Machine Learning (MCML), Germany \\
  \small
  $^6$Univ.\ Grenoble Alpes, Inria, CNRS, Grenoble INP, LJK, France \\
}

\begin{document}

\maketitle

% abstract and visual summary
\begin{abstract}
Earth observation (EO) foundation models (FMs) are increasingly trained on multisensor data, spanning multispectral imagery (MSI), synthetic aperture radar (SAR), and derived geospatial layers, but hyperspectral imagery (HSI) remains underrepresented. Conversely, existing hyperspectral FMs are trained on HSI alone, leaving joint pretraining and fusion of HSI with co-located EO sensors unexplored.

We introduce \modelname, a hierarchical transformer for multisensor EO input with heterogeneous spectral dimensionality. The architecture combines spectral tokenization for hyperspectral inputs, sensor-specific encoders, a cross-sensor fusion module, and a shared hierarchical encoder, enabling joint processing of HSI and lower-channel observations. To pretrain \modelname, we curate \datasetname, a dataset that co-locates HSI from three spaceborne sensors (\EnMAP, \EMIT, \DESIS) with \Stwo, \Landsat optical imagery, Landsat land surface temperature (LST), and \Sone SAR, over common geographic footprints. It comprises approximately 2M globally distributed locations, 25M georeferenced patches, and over 40TB of data. Pretraining uses a Joint-Embedding Predictive Architecture (JEPA)-style objective that matches representations between global views and single-sensor local views from the same location.
We evaluate \modelname on hyperspectral downstream tasks and standard EO benchmarks following the PANGAEA protocol, achieving state-of-the-art results across both evaluation settings.

\end{abstract}
\begin{figure*}[th!]
\centering
\includegraphics[width=\textwidth]{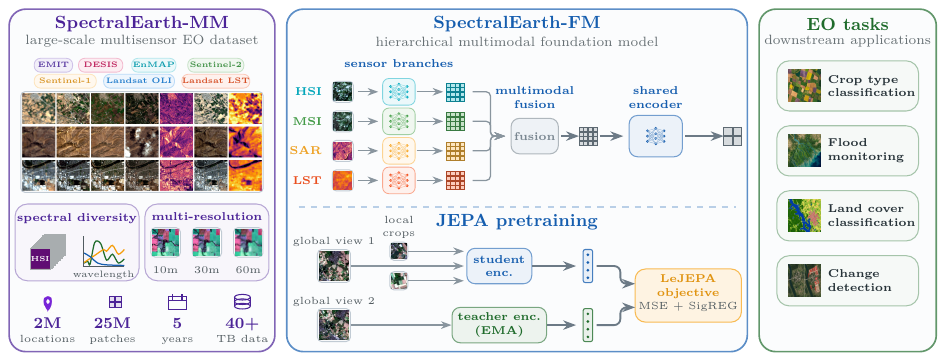}
\caption{Overview of \modelname architecture and pretraining
objective (center), trained on a multi-modal data corpus \datasetname (left) and evaluated on EO benchmarks (right).}
\label{fig:overview}
\end{figure*}

% main article
\section{Introduction}
\label{sec:introduction}
Earth observation (EO) foundation models (FMs) are increasingly trained on a plurality of sensors and geospatial layers, including  multispectral imagery (MSI, with channel bandwidth $\sim100nm$), synthetic aperture radar (SAR), digital elevation models (DEM), vegetation indices (e.g., NDVI), and land use land cover (LULC) maps \citep{fuller2023croma,jakubik2023prithvi,astruc2025anysat,jakubik2025terramind,wang2025copernicusfm}.
In parallel, recent EO models on hyperspectral imagery (HSI, with channel bandwidth $\sim10nm$) have demonstrated that large-scale pretraining can learn transferable spectral--spatial representations for tasks requiring fine-grained spectral discrimination \citep{wang2024hypersigma,braham2024spectralearth,baumann2025carl}.
However, the integration of HSI into multi-sensor EO FMs remains underexplored, as previous efforts related to HSI pretraining have not included spatially aligned MSI, SAR, and thermal imagery.

The lack of HSI in EO FMs also reflects differences in modeling: HSI contains high-dimensional, spectrally contiguous measurements with strong inter-band correlations and sensor-dependent noise.
As a result, HSI models typically employ explicit spectral--spatial processing, such as spectral grouping or band-aware tokenization \citep{hong2022spectralformer,wang2024hypersigma,braham2024spectralearth}. A deep learning architecture that combines HSI with MSI, SAR, and land surface temperature (LST) should preserve spectral processing while also supporting broadband optical inputs~\citep{leonardi2026spectral}.

In this work, we present \modelname, a spatial-spectral EO foundation model built on a hierarchical transformer architecture, specifically designed for spectrally diverse EO inputs. Its architecture combines a spectral tokenizer for HSI, sensor-specific branches, a multisensor fusion module, and a shared hierarchical encoder. We pretrain \modelname with a Joint-Embedding Predictive Architecture (JEPA)-style objective adapted for multimodal EO fusion. Instead of reconstructing raw pixels or sensor measurements, the objective matches representations between global views and sensor-dropped local crops from the same location. This design avoids direct reconstruction across sensors with different measurement characteristics to (a) reduce the impact of sensor-specific noise and to (b) prevent a few spectral channels from serving as the basis for predicting hundreds of others.
To support pretraining at scale, we introduce \datasetname, a multi-sensor dataset constructed from the EMIT \citep{green2020earth}, EnMAP \citep{guanter2015enmap}, and DESIS \citep{krutz2019instrument} spaceborne HSI missions and spatially aligned with \Stwo, \Landsat MSI (OLI), \Landsat LST, and \Sone SAR. Comprising two million globally balanced geographic locations and over 40TB of data, \datasetname provides 25 million co-located multimodal samples with multitemporal coverage where available.

We evaluate \modelname on hyperspectral downstream tasks and standard EO benchmarks under the PANGAEA protocol \citep{marsocci2024pangaea}, and demonstrate generalization to unseen sensors via simple spectral interpolation to a pretraining sensor, achieving competitive performance against recent foundation models~\citep{xiong2024dofa,waldmann2025panopticon,baumann2025carl,jakubik2025terramind}.

This paper makes the following contributions, illustrated in \cref{fig:overview}:
\begin{itemize}
    \item \textbf{\datasetname}, a globally balanced, multitemporal corpus of over 40TB spanning HSI, MSI, SAR, and LST, centered on spaceborne hyperspectral missions (\cref{sec:spectralearth-mm}).
    \item \textbf{\modelname}, an EO foundation model based on a hierarchical vision transformer with dedicated spatial--spectral processing for multisensor optical and radar fusion (\cref{sec:architecture}).
    \item A \textbf{JEPA-style pretraining strategy} tailored to heterogeneous multisensor data (\cref{sec:objective}), with evaluation on ten hyperspectral downstream tasks and community established PANGAEA benchmarks (\cref{sec:experiments}).
\end{itemize}

\section{Related work}
\label{sec:related_work}

\paragraph{EO foundation models}
Early EO self-supervised learning~\citep{wang2022self} focused on single optical sensors, using contrastive objectives based on seasonal, geographic, or temporal structure \citep{ayush2021iccv-geographyaware,manas2021seasonalcontrast}. Later works shifted to masked reconstruction at scale: SatMAE targeted temporal and multispectral satellite imagery \citep{cong2022satmae}, Prithvi(-2) applied the same paradigm to the harmonized Landsat-Sentinel product (HLS) \citep{jakubik2023prithvi,szwarcman2024prithvi2}, and SpectralGPT extended it to Sentinel-2 via spectral--spatial tokenization~\citep{hong2024spectralgpt}.
A second line of work moves beyond single optical inputs to multiple sensors and geospatial products. While CROMA pairs \Sone/\Stwo observations \citep{fuller2023croma}, RingMo(-Sense) brings masked image modeling to multisensor and spatiotemporal settings \citep{sun2022ringmo,yao2023ringmo}. Rather than a shared encoder, OmniSat employs sensor-specific encoders with resolution-aware tokenization \citep{astruc2024omnisat}, and AnySat extends this architecture further with JEPA-style pretraining \citep{astruc2025anysat}. Beyond purely observational inputs, MMEarth incorporates derived products and auxiliary pretext tasks \citep{nedungadi2024mmearth}, while TerraMind learns cross-modal prediction across optical, radar, and geospatial layers \citep{jakubik2025terramind}. TerraFM trains on Sentinel-1/2 with modality augmentation and cross-attention fusion \citep{danish2025terrafm}.
Unlike prior multisensor EO FMs, which exclude HSI, we pair multiple spaceborne HSI sensors with MSI, SAR, and LST and develop an architecture that explicitly handles high spectral dimensionality.

\paragraph{Hyperspectral foundation models}
The high spectral dimensionality of HSI motivates explicit spectral--spatial processing in EO FM architectures. Some works adapt masked autoencoders to HSI using blockwise patch embeddings, spectral positional embeddings, and spatial--spectral attention \citep{scheibenreif2023cvprw-masked}. HyperSIGMA \citep{wang2024hypersigma} uses separate spatial and spectral transformers with cross‑attention fusion to pretrain on Hyperion EO-1 \citep{pearlman2003hyperion} and Gaofen-5 imagery \citep{liu2019advanced}. SpectralEarth \citep{braham2024spectralearth} uses convolutional spectral adapters with standard vision backbones, with several self-supervised objectives, and provides an EnMAP-based pretraining dataset and downstream HSI benchmarks.
Pixel-level spectral models such as HyperSL \citep{kong2025hypersl} instead learn representations from spectra directly, without explicitly modeling spatial context. Further work explores continuous spectral encoders through implicit neural representations that treat wavelength as a continuous variable~\citep{chen2023spectral}. HyperFM uses spectral grouping with lightweight group‑wise and cross‑group attention to model hyperspectral structure efficiently \citep{tushar2026hyperfM}.
These works focus on HSI inputs with explicit spectral modeling. Our setting combines HSI with additional EO modalities within a unified pretraining framework.

\paragraph{Any-sensor foundation models}
A separate line of work achieves sensor generalization by conditioning patch embeddings on wavelength or sensor metadata, rather than learning sensor-specific parameters. DOFA generates embeddings from wavelength metadata via a dynamic hypernetwork \citep{xiong2024dofa}; Copernicus-FM extends this to non-spectral Sentinel modalities \citep{wang2025copernicusfm}. Panopticon trains across varying spectral configurations using channel sub-sampling and early band aggregation \citep{waldmann2025panopticon}.
SpecAware conditions token encoding jointly on sensor meta-attributes and image content, targeting multisensor HSI \citep{ji2025specaware}. CARL learns camera-agnostic representations across RGB, MSI, and HSI via wavelength-aware spectral encoding \citep{baumann2025carl}, and SMARTIES map heterogeneous MSI bands into a shared spectrum-aware space \citep{sumbul2025smarties}. These approaches prioritize generalization to arbitrary sensors at the cost of sensor-specific spectral modeling. In contrast, \modelname learns dedicated branches with explicit spectral tokenization for each pretraining sensor, and transfers to unseen sensors by spectral interpolation to the nearest trained branch.
\section{\datasetname dataset}
\label{sec:spectralearth-mm}
\datasetname provides a large-scale multisensor EO dataset centered on spaceborne hyperspectral imagery used to pretrain \modelname.

\subsection{Dataset overview}
\datasetname uses spaceborne HSI as the reference observation. It includes HSI from three sensors: \EnMAP, \EMIT, and \DESIS. These sensors provide complementary hyperspectral coverage: \DESIS observes the visible and near-infrared (VNIR) range at high spectral resolution, while \EnMAP and \EMIT cover VNIR and shortwave infrared (SWIR) wavelengths, with different spatial resolutions and acquisition geometries. For each selected HSI footprint, we pair additional EO sensors from \Stwo, \Landsat optical imagery, Landsat LST, and \Sone.

\begin{figure}[t!]
  \centering
  \includegraphics[width=\linewidth]{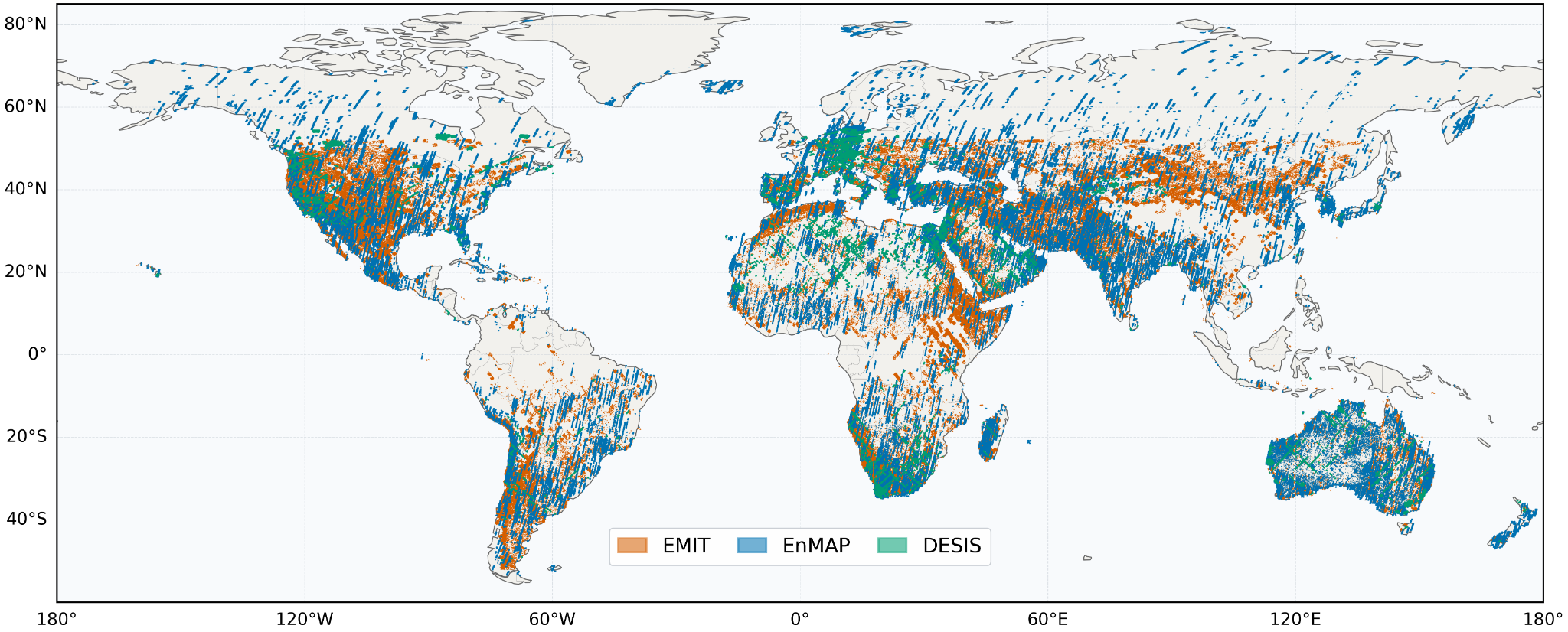}
  \caption{\textit{Spatial coverage of \datasetname.} Global distribution of HSI anchor patches in \datasetname. Colors indicate the HSI sensor associated with each acquisition. }
  \label{fig:dataset_global_coverage}
\end{figure}

All observations are organized over a common geographic footprint of $(3.84\,\mathrm{km})^2$. This footprint corresponds to $128^2$ pixels for \EnMAP, \DESIS and \Landsat at 30 m resolution, $64^2$ pixels for \EMIT at 60 m resolution, and $384^2$ pixels for \Sone and \Stwo at 10 m resolution. The resulting dataset spans approximately 2M globally distributed locations, 25M georeferenced patches, and over 40 TB of data. Multiple HSI acquisition dates are retained for the same geolocation when available, although the temporal density varies across regions and sensors. \Cref{fig:dataset_global_coverage} shows the global distribution of the selected HSI anchor footprints, covering approximately 20\% of the global land surface.

\begin{table}[t!]
  {\centering
  \caption{
  \textit{Representative EO pretraining datasets.}
  We group datasets into general multi-sensor remote sensing and HSI-specific pre-training datasets. ``Spat. align'' indicates whether multiple observation types are provided over common geographic footprints. ``Multi-temp.'' indicates the availability of multiple acquisition dates for the same geolocation.
  }
  \label{tab:pretrain_datasets}
    \footnotesize\textit{legend:} \cmark\dots available, \omark\dots partially available, \xmark\dots unavailable\\
  \resizebox{\linewidth}{!}{%
  \begin{tabular}{lcccccccccc}
    \toprule
    Dataset & HSI & MSI & SAR & LST & Spat. align & Multi-temp. & Size [TB] & Loc. [M] & Patches [M] & Sensors$^\dagger$ \\
    \midrule
    SatlasPretrain \citep{bastani2023satlaspretrain} & \xmark & \cmark & \xmark & \xmark & \omark & \cmark & 30.0 & 0.86 & $\sim$9.0\phantom{0} & 2 \\ % S2, NAIP; https://satlas-pretrain.allen.ai
    SSL4EO-S12 \citep{wang2023ssl4eo} &\xmark & \cmark & \cmark & \xmark & \cmark & \cmark & \phantom{0}3.7 & 0.25 & \phantom{0}1.5 & 2\\ % S1, S2; https://github.com/zhu-xlab/SSL4EO-S12
    TerraMesh \citep{blumenstiel2025terramesh}& \xmark & \cmark & \cmark & \xmark & \cmark & \xmark & 17.0 & 9.00 & 64.0 & 3 \\ % S1, S2, DEM, LULC=S1+S2, NDVI=S2; https://huggingface.co/datasets/ibm-esa-geospatial/TerraMesh
    MMEarth \citep{nedungadi2024mmearth}& \xmark & \cmark & \cmark & \xmark & \cmark & \xmark & \phantom{0}0.6 & 1.20 & \phantom{0}7.2 & 4 \\ % S1, S2, ASTER, ETH canopy (GEDI); https://vishalned.github.io/mmearth
    MajorTOM-Core \citep{francis2024majortom}& \xmark & \cmark & \cmark & \xmark & \cmark & \xmark & 62.0 & 1.47 & \phantom{0}6.0 & 4 \\ % S1, S2, VIIRS, DEM; https://huggingface.co/Major-TOM
    Copernicus-Pretrain \citep{wang2025copernicusfm}& \xmark & \cmark & \cmark & \xmark & \cmark & \cmark & 20.7 & 0.31 & 18.7 & 5 \\ % S1, S2, S3, S5P, DEM; https://github.com/zhu-xlab/Copernicus-FM#copernicus-pretrain
    \midrule
    HySpecNet-11k \citep{fuchs2023hyspecnet}& \cmark & \xmark & \xmark & \xmark & \xmark & \xmark & \phantom{0}0.1 & 0.01 & 0.01 & 1 \\ % EnMAP; https://hyspecnet.rsim.berlin/
    HyperFM250k \citep{tushar2026hyperfM} & \cmark & \xmark & \xmark & \xmark & \xmark & \xmark & \phantom{0}3.0 & 0.25 & 0.25 & 1 \\ % NASA PACE; https://github.com/umbc-sanjaylab/HyperFM
    Hyper-400K \citep{ji2025specaware}& \cmark & \xmark & \xmark & \xmark & \xmark & \xmark & 17.0 & 0.40 & 0.40 & 1 \\ % AVIRIS; https://github.com/busbyjrj/SpecAware#hyper-400k
    HyperGlobal-450K \citep{wang2024hypersigma}& \cmark & \xmark & \xmark & \xmark & \xmark & \xmark & \phantom{0}0.4 & 0.45 & 0.45 & 2 \\ % Hyperion and Gaofen-5B % https://github.com/WHU-Sigma/HyperSIGMA    
    SpectralEarth \citep{braham2024spectralearth}& \cmark & \xmark & \xmark & \xmark & \xmark & \cmark & \phantom{0}3.3 & 0.42 & 0.54 & 1 \\ % EnMAP; https://github.com/AABNassim/spectral_earth
    \midrule
    \textbf{\datasetname} (ours) & \cmark & \cmark & \cmark & \cmark & \cmark & \cmark & 40.0 & 2.00 & 25.00 & 7 \\ % EnMAP, EMIT, DESIS, L8/9, L8 LST, S1, S2
    \bottomrule
  \end{tabular}
  }
  }
  \relsize{-2}$\dagger$: derived products from the same instrument count as one; geospatial layers and non-EO data such as climate and weather data excluded
  \vspace{-2ex}
\end{table}

\Cref{tab:pretrain_datasets} compares \datasetname with existing EO pretraining datasets. Existing multisensor datasets provide large-scale optical and radar observations, sometimes with derived geospatial layers, but generally do not include spaceborne HSI. Conversely, HSI datasets provide dense spectral measurements but are typically not paired with common EO sensors over the same locations. \datasetname addresses both by pairing spaceborne HSI with multispectral, radar, and LST observations at scale.

\subsection{Dataset construction}

\paragraph{HSI preprocessing and footprint extraction}
We start from approximately 60k \EnMAP, 33k \EMIT, and 11k \DESIS tiles collected from mission archives. We apply sensor-specific quality control, including georeferencing checks, cloud screening, and removal of noisy spectral bands. The remaining scenes are patchified into non-overlapping $(3.84\,\mathrm{km})^2$ footprints. Patches with missing data are discarded, and the remaining samples are grouped by geographic footprint and timestamp.

\paragraph{Rebalancing HSI locations}
The initially assembled HSI archive is highly imbalanced across sensors and geographic regions. 
The raw collection is dominated by \EMIT-only and \EMIT--\EnMAP sensor combinations, with \DESIS and \EMIT coverage being restricted by the flight path of the International Space Station (ISS). We therefore rebalance the HSI anchor patches before multisensor pairing. For each location, we retrieve annual AlphaEarth embeddings \citep{brown2025alphaearth}, cluster them hierarchically, and subsample overrepresented configurations within the resulting clusters, following~\cite{voautomatic}. This reduces spatial and semantic redundancy while preserving scarce sensor configurations and retaining all timestamps for selected locations. After rebalancing, the HSI archive contains approximately 1.4M \EnMAP locations, 1.4M \EMIT locations, and 275k \DESIS locations. These counts are not additive because a single geographic location may contain observations from multiple HSI sensors and dates.

\paragraph{Multisensor pairing and temporal filtering}
We pair the selected HSI footprints with additional observations, namely \Stwo, \Landsat MSI, Landsat LST, and \Sone SAR. Optical observations are filtered for low cloud coverage (<10\%). For each HSI timestamp, these additional observations are selected within a temporal proximity window to ensure cross-sensor coherence. After pairing, we apply temporal filtering to reduce redundant acquisitions while preserving multiple dates when available. Additional details can be found in \Cref{app:dataset_details}. 
\section{\modelname}
\label{sec:method}

The sensors in \datasetname differ substantially in spectral dimensionality
and spatial resolution, motivating three design choices. First, HSI requires explicit spectral modeling before spatial encoding: a single patch projection would mix hundreds of bands before any spectral structure is captured. Second, patch sizes are chosen per sensor, so each token covers the same ground extent, producing a common spatial token layout without resampling. Third, a hierarchical encoder captures local structure progressively, supporting fine-grained dense prediction. \modelname addresses these through sensor-specific input branches, a cross-sensor fusion module, and a shared hierarchical encoder, and supports single- and multisensor inference over sensors seen during pretraining, as illustrated in \cref{fig:spectraljepa_architecture}.

\subsection{Network architecture}
\label{sec:architecture}

\begin{figure*}[t]
  \centering
  \includegraphics[width=\textwidth]{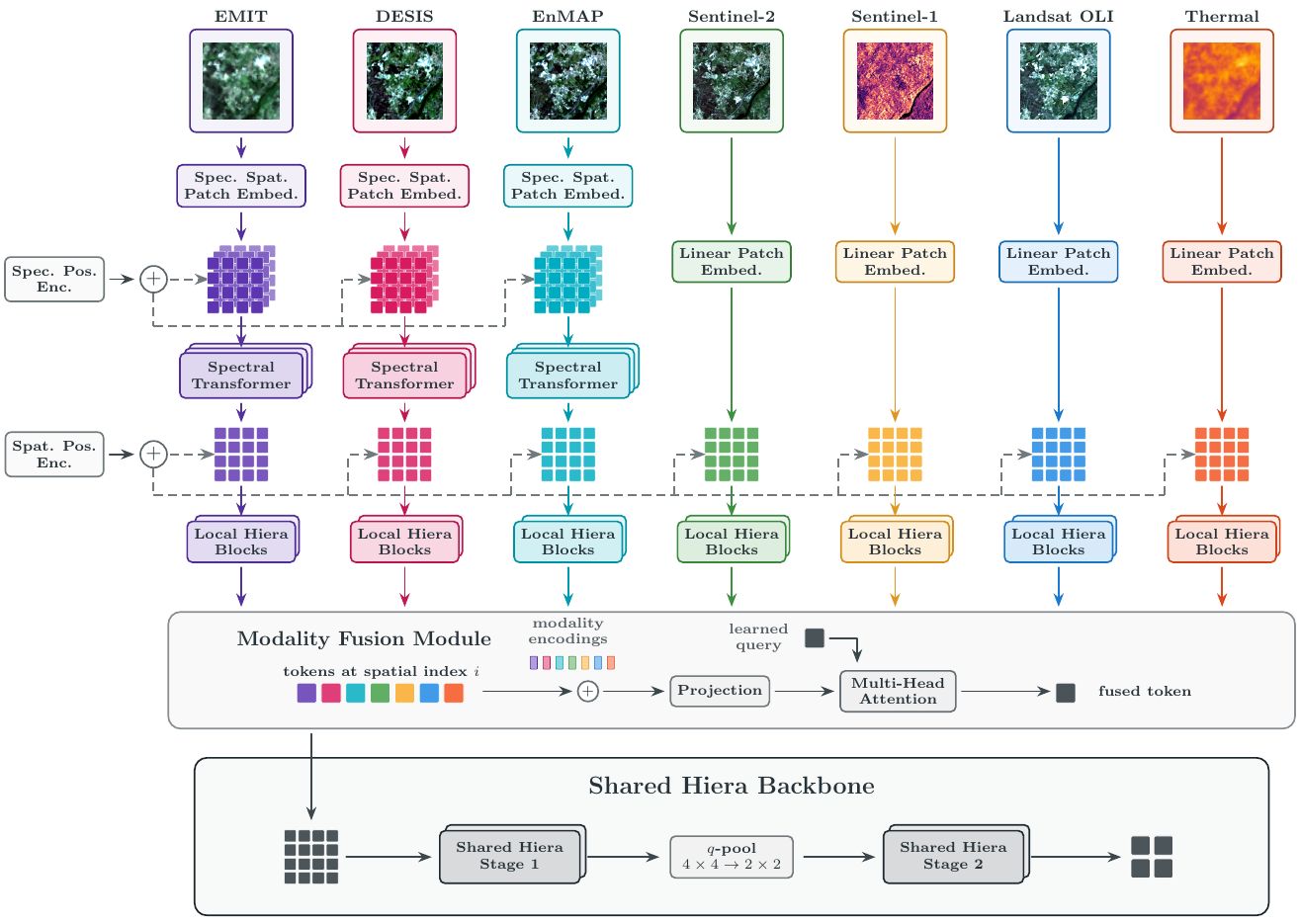}
  \caption{
  \textit{\modelname architecture.}
  Each available input is mapped to a common spatial token grid.
  HSI inputs use spectral tokenization before spatial encoding, while lower-dimensional sensors use linear projections.
  Local hierarchical branches process sensor-specific features before cross-modal fusion.
  The fused tokens are passed to a shared hierarchical backbone.
  }
  \label{fig:spectraljepa_architecture}
\end{figure*}

\paragraph{Input data}
Let $\mathcal{S}$ denote the set of pretraining sensors. For a given
geographic location, let $\mathcal{A} \subseteq \mathcal{S}$ be the set of
available inputs. Each input $x_s$ has $C_s$ spectral channels and sensor-specific spatial dimensions.
Patch sizes are set per sensor so that all branches produce the same
spatial token layout.

\paragraph{Spectral tokenization}
HSI inputs are processed by a sensor-specific spectral tokenizer.
The $C_s$ bands within each spatial patch are partitioned into $G_s$ groups
of contiguous spectral bands. Each group is embedded by a dedicated patch
projection $E_s^{\mathrm{spec}}$, producing a sequence of $G_s$ tokens per
spatial position. Learnable spectral position encodings $P_\lambda$ are
added to inject spectral order information, and a spectral transformer
$\mathrm{SpecTr}$ processes the resulting sequence. The group tokens are
then aggregated into a single spatial token by a learned query attending
over the projected group tokens ($\mathrm{Agg}$):
\begin{equation}
    T_s = \mathrm{Agg}\!\left(
        \mathrm{SpecTr}(E_s^{\mathrm{spec}}(x_s) + P_\lambda)
    \right) \in \mathbb{R}^{L \times D},
\end{equation}
where $L$ is the number of spatial patches and $D$ is the embedding
dimension. All other sensors (\Stwo, \Landsat optical, \Sone, Landsat LST)
use a standard convolutional patch projection $E_s^{\mathrm{patch}}$:
\begin{equation}
    T_s = E_s^{\mathrm{patch}}(x_s) \in \mathbb{R}^{L \times D}.
\end{equation}

\paragraph{Sensor-specific local branches}
After tokenization, shared spatial position embeddings $P_p$ are added and
sensor-specific local hierarchical blocks $\mathrm{LHB}_s$ process the
tokens to produce branch embeddings:
\begin{equation}
    H_s = \mathrm{LHB}_s(T_s + P_p) \in \mathbb{R}^{L_\ell \times D_\ell}.
\end{equation}
where $L_\ell \leq L$ is the pooled spatial token count and $D_\ell$ is the branch output width, which also serves as the input width of the shared encoder.
The local blocks follow a Hiera-style design~\citep{ryali2023hiera}, using
window-based attention in early stages before pooling to coarser
representations. Each branch operates independently, handling
sensor-specific input statistics without cross-sensor interaction.

\paragraph{Cross-sensor fusion}
After local encoding, the branch embeddings $\{H_s : s \in \mathcal{A}\}$
are fused at each spatial position. Each branch embedding is first projected
to a fusion dimension $D_f$ and augmented with a learned sensor-specific embedding.
A learned query then attends over the resulting sensor tokens, and the output
is mapped back to dimension $D_\ell$, producing a single fused token grid $F \in \mathbb{R}^{L_\ell \times D_\ell}$.
The fusion module reuses the projected-attention design of the spectral tokenizer, replacing spectral groups with sensor branches.

\paragraph{Shared hierarchical encoder}
The fused token grid $F$ $\in \mathbb{R}^{L_\ell \times D_\ell}$ is passed to a shared Hiera
encoder~\citep{ryali2023hiera}, producing a representation
$Z$. For classification, the final tokens are mean-pooled; for dense
prediction, intermediate feature maps from each hierarchical stage can
be passed to task-specific decoders.

The full pipeline from inputs to representation, visualized in \cref{fig:spectraljepa_architecture},  is:
\begin{equation}
    \{x_s\}_{s \in \mathcal{A}}
    \;\xrightarrow{\text{tokenize}}\; \{T_s\}
    \;\xrightarrow{\text{local}}\; \{H_s\}
    \;\xrightarrow{\text{fuse}}\; F
    \;\xrightarrow{\text{shared}}\; Z.
\end{equation}

\subsection{Multisensor pretraining}
\label{sec:objective}

\begin{figure*}[t]
  \centering
  \includegraphics[width=\textwidth]{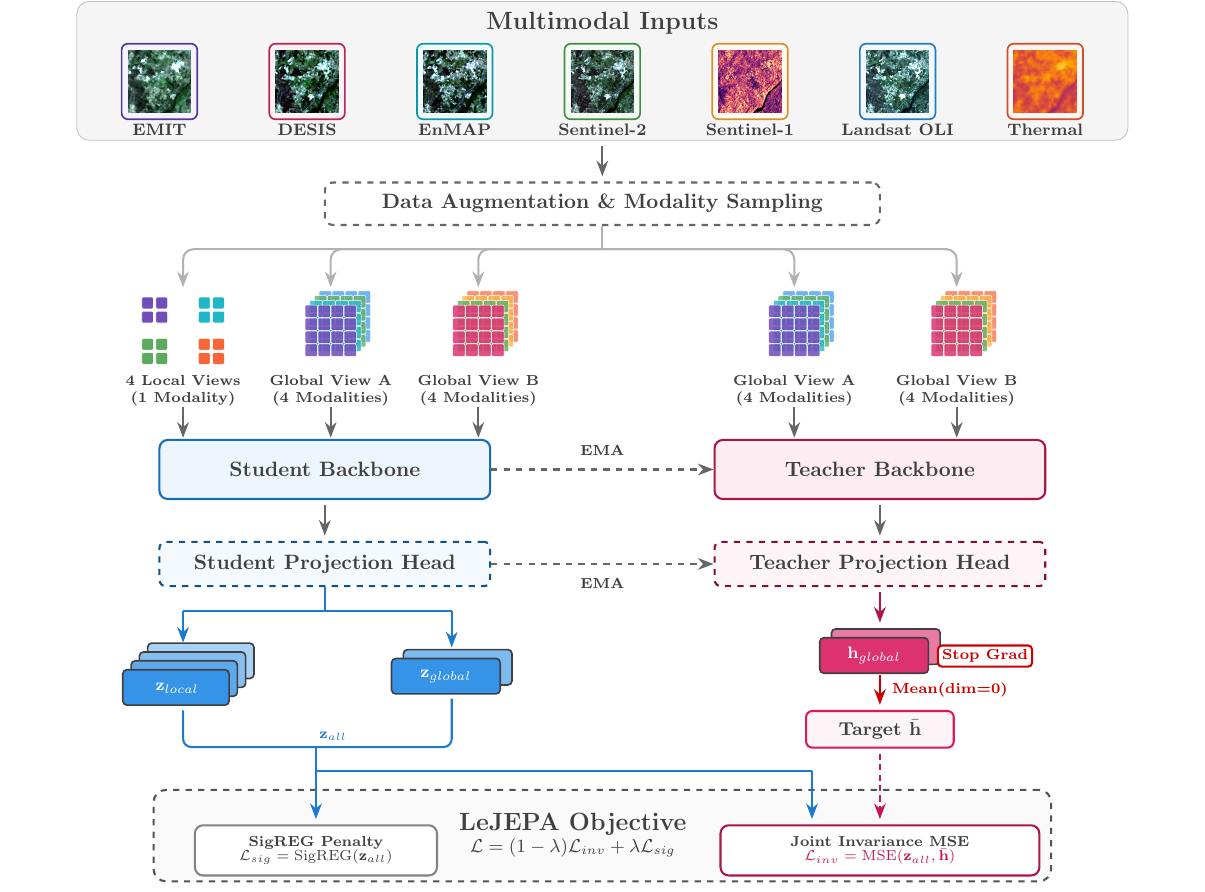}
  \caption{
  \textit{\modelname pretraining.}
  Global teacher views define a stop-gradient latent target.
  The student processes global, local, and sensor-dropped views from the same geographic location and predicts the teacher target.
  SIGReg is applied to the stacked student projections.
  }
  \label{fig:spectraljepa_objective}
\end{figure*}

\paragraph{Objective design}
Pixel-level reconstruction~\citep{he2022masked} is a challenging pretraining target: HSI measurements are affected by atmospheric conditions, calibration, and sensor noise, and these challenges multiply across modalities that differ in spectral dimensionality, radiometric scale, and noise statistics. Instead, we adopt a latent prediction objective~\citep{assran2023ijepa, balestriero2025lejepa}, where the model predicts latent representations rather than pixel measurements, focusing on learning on semantic structure over low-level sensor characteristics.

\paragraph{View construction}
Each location in \datasetname provides between five and seven modalities
depending on HSI sensor coverage. We sample $V$ global views and $N$ local
views per location. Global views are spatial crops containing
four randomly sampled modalities, with consistent spatial transforms
across modalities. Local views are smaller spatial crops that retain a single modality each.

\paragraph{Teacher--student objective}
Learning joint and single-sensor representations simultaneously is difficult with a symmetric objective: the model can satisfy the invariance loss by collapsing toward weak representations that are easy to predict from partial views. We address this with an asymmetric teacher--student design~\citep{grill2020bootstrap, chen2021exploring}: a teacher network maintained as an exponential moving average (EMA) of the student processes only the global views and defines a stable prediction target; the student processes all views and is trained to match it.

Let $f_\theta$ and $p_\phi$ denote the student encoder and projector, and $f_{\bar{\theta}}$, $p_{\bar{\phi}}$ their EMA counterparts. The teacher target for sample $b$ is the stop-gradient mean of teacher projections over the $V$ global views:
\begin{equation}
    \bar{h}_b =
    \mathrm{sg}\!\left(
        \frac{1}{V}\sum_{v=1}^{V}
        p_{\bar{\phi}}\!\left(f_{\bar{\theta}}(v^g_b)\right)
    \right).
\end{equation}
The student produces $z_{v,b} = p_\phi(f_\theta(v_b))$ for each view, and
the invariance loss follows~\citep{balestriero2025lejepa}:
\begin{equation}
    \mathcal{L}_{\mathrm{inv}}
    = \frac{1}{B\,|\mathcal{V}|}
    \sum_{b=1}^{B}
    \sum_{v \in \mathcal{V}}
    \left\lVert z_{v,b} - \bar{h}_b \right\rVert_2^2,
\end{equation}
where $\mathcal{V}$ denotes all views, global and local, and $B$ is the
batch size. Predicting the fused global representation from a single modality
aligns single-sensor representations with the joint representation space. 
To prevent
collapse~\citep{balestriero2025lejepa}, SIGReg is applied to the stacked
student projections across all views and samples. The final objective is:
\begin{equation}
    \mathcal{L}
    = (1 - \lambda)\,\mathcal{L}_{\mathrm{inv}}
    + \lambda\,\mathcal{L}_{\mathrm{SIGReg}}
      \!\left(\{z_{v,b}\}_{v,b}\right).
\end{equation}
where $\lambda \in [0,1]$ controls the regularization weight.

\section{Experiments}
\label{sec:experiments}

\subsection{Experimental setup}
\label{sec:setup}

\paragraph{Pretraining}
We pretrain \modelname on \datasetname following the objective described in \Cref{sec:method}. We sample two global views per location, each containing four randomly drawn modalities, and four local views each with a single modality. We set the SIGReg weight to $\lambda = 0.015$ and the projection
dimension to $128$. Training runs for $50$ epochs with AdamW, a cosine
learning rate schedule with linear warmup, a peak learning rate of
$10^{-5}$, and a batch size of $240$ on 48 NVIDIA A100 GPUs. More
implementation details can be found in \Cref{app:model_details,app:training_details}.

\paragraph{Downstream evaluation}
We evaluate \modelname on hyperspectral benchmarks 
(\Cref{sec:hsi_eval}) and general EO benchmarks under the PANGAEA 
protocol (\Cref{sec:pangaea}). All evaluations use a frozen encoder 
and report mean intersection over union (mIoU). 

\paragraph{Sensor branch routing}
A downstream input can be routed to one or more pretrained sensor 
branches. In the \emph{single-branch} setting (\modelname-s), inputs 
are routed to the branch with the closest spectral coverage and spatial 
resolution (e.g., EO-1~Hyperion $\to$ EnMAP, GF-5 $\to$ EnMAP). In the \emph{multi-branch} setting (\modelname-a), the same 
acquisition is passed to all pretrained optical branches, and the 
resulting features are concatenated before the downstream head. HSI inputs are remapped by wavelength interpolation (HSI branches) or band averaging (MSI branches). For PANGAEA benchmarks, \modelname-s routes to the \Stwo branch and \modelname-a routes to the \Stwo and OLI branches. Both models use \Sone data when available. %The multi-branch setting requires only a single acquisition and differs from the fusion experiments in  \Cref{sec:real_fusion}.

%% -------------------------------------------------------
\subsection{Hyperspectral benchmarks}
\label{sec:hsi_eval}

We evaluate on ten benchmarks: eight from the SpectralEarth suite
\citep{braham2024spectralearth}, covering EnMAP (CDL, BD-Foret, BNTD,
EuroCrops, NLCD, TreeMap), DESIS (DESIS-CDL), and EO-1 Hyperion (EO-1 CDL),
plus GF-5 Wuhan \citep{hong2023cross} and OxHyperMinerals
\citep{ruzicka2025hyperspectralvits}---for the latter, we use EMIT Level-2A reflectance instead of the original Level-1 product  and a $10\%$ data split. We compare against CARL \citep{baumann2025carl}, DOFA-B/L \citep{xiong2024dofa},
HyperSIGMA-B/L \citep{wang2024hypersigma}, Panopticon \citep{waldmann2025panopticon}, SpecAware \citep{ji2025specaware}, and Spectral ViT-B/L
\citep{braham2024spectralearth}. All models use a SegFormer-style
segmentation head~\citep{xie2021segformer} 
and a learning rate of $10^{-4}$.

\begin{table*}[t]
\centering
\caption{Performance of hyperspectral models on downstream tasks. Scores are mIoU (\%); best is \textbf{bold} and second best is \underline{underlined}.}
\label{tab:hsi-main}
\vspace{0.25em}
\footnotesize
\setlength{\tabcolsep}{3pt}
\begin{tabular}{lccccccccccc}
\toprule
Model & CDL & BD-F & BNTD & EuCrops & NLCD & TreeMap & D-CDL & EO1-CDL & GF5-WH & OxHyp & Rank \\
\midrule
CARL & 57.4 & 52.3 & 36.3 & 47.0 & 34.3 & 35.6 & 58.5 & 52.0 & 40.4 & 61.1 & 7.3 \\
DOFA-B & 55.6 & 47.1 & 35.6 & 44.9 & 32.7 & 33.8 & 55.2 & 55.6 & 41.9 & 61.2 & 8.3 \\
DOFA-L & 54.6 & 45.1 & 35.4 & 44.4 & 32.1 & 32.2 & 55.6 & 53.7 & 41.5 & 56.2 & 9.2 \\
HyperSIGMA-B & 58.2 & 52.4 & 35.9 & 47.6 & 34.6 & 35.2 & 49.4 & 57.7 & 40.4 & 65.6 & 6.8 \\
HyperSIGMA-L & 58.4 & 50.0 & 35.3 & 46.0 & 33.9 & 34.6 & 48.2 & 57.3 & 40.3 & 64.4 & 8.1 \\
Panopticon & 61.5 & 60.4 & 38.5 & 49.6 & 38.6 & 38.4 & 57.7 & 63.8 & \textbf{44.1} & 62.6 & 4.9 \\
SpecAware & 51.9 & 43.5 & 33.3 & 44.2 & 31.7 & 29.6 & 46.8 & 50.8 & 36.1 & 37.9 & 11.0 \\
Spec. ViT-B & 70.9 & 67.2 & 41.3 & 60.7 & 41.1 & 39.0 & 62.4 & 64.0 & 43.8 & 66.1 & 3.7 \\
Spec. ViT-L & 70.5 & \underline{69.7} & 42.0 & 60.9 & \underline{42.9} & \underline{41.3} & 64.0 & 64.5 & \underline{43.9} & 65.8 & 2.8 \\
\midrule
SpectralEarth-FM-s & \underline{72.4} & 69.3 & \underline{44.0} & \underline{62.1} & 42.6 & 41.3 & \underline{65.5} & \textbf{68.7} & 42.9 & \underline{77.6} & \underline{2.5} \\
SpectralEarth-FM-a & \textbf{74.0} & \textbf{70.9} & \textbf{44.9} & \textbf{63.5} & \textbf{43.7} & \textbf{42.1} & \textbf{66.3} & \underline{68.0} & 43.2 & \textbf{78.9} & \textbf{1.4} \\
\bottomrule
\end{tabular}
\end{table*}

\paragraph{Results}
\modelname-s achieves an average rank of $2.5$ and \modelname-a achieves $1.4$, ranking first overall. On the six EnMAP benchmarks, \modelname-a outperforms all baselines, including Spec.~ViT-L. \modelname also outperforms any-sensor models trained on EnMAP data such as Panopticon, CARL, and DOFA. On EO-1 CDL, \modelname achieves the highest score among all evaluated models, suggesting that spectral interpolation to the EnMAP branch can transfer to sensors with overlapping VNIR--SWIR coverage. On GF-5 Wuhan, both variants outperform HyperSIGMA-B and HyperSIGMA-L by $2.5$--$2.9$~mIoU despite HyperSIGMA being pretrained on Gaofen-5 imagery; Panopticon performs best on this task. On OxHyperMinerals, both variants substantially outscore all baselines. Multi-branch routing improves over single-branch routing on nine of ten benchmarks, with the largest gains on CDL ($+1.6$), BD-Forêt ($+1.6$), and EuroCrops ($+1.4$), suggesting that the features extracted by the different sensor branches encode complementary information.

\paragraph{Sensor branch routing ablation}
\Cref{tab:branch-ablation} isolates the effect of branch selection for
\modelname. Among single-branch routes, EnMAP produces the highest scores on nine of ten benchmarks; EMIT leads on OxHyperMinerals ($77.2$), where it is the source sensor. DESIS scores below EnMAP and EMIT on most benchmarks but leads on DESIS-CDL, where it is the in-distribution sensor for that task. The Landsat optical branch scores lowest on most benchmarks, reflecting its coarser spectral sampling relative to the HSI and S2 branches. Combining all three HSI branches improves over the best single-branch result on most benchmarks. Adding S2 and OLI yields modest gains on seven out of ten tasks, indicating that each branch contributes complementary features.

\begin{table*}[t]
\centering
\caption{Branch-mapping ablation for SpectralEarth-FM. Scores are mIoU (\%); best is \textbf{bold} and second best is \underline{underlined}.}
\label{tab:branch-ablation}
\vspace{0.25em}
\footnotesize
\setlength{\tabcolsep}{3pt}
\begin{tabular}{lcccccccccc}
\toprule
Target branches & CDL & BD-F & BNTD & EuCrops & NLCD & TreeMap & D-CDL & EO1-CDL & GF5-WH & OxHyp \\
\midrule
EnMAP & 72.3 & 69.6 & 43.9 & 62.3 & 42.5 & 41.4 & 62.0 & \underline{69.0} & 42.7 & 76.0 \\
DESIS & 69.5 & 63.2 & 42.6 & 57.7 & 40.2 & 39.7 & 64.8 & 62.6 & 41.3 & 65.5 \\
EMIT & 70.7 & 68.4 & 42.8 & 60.0 & 41.1 & 41.1 & 62.3 & 67.8 & 42.6 & 77.2 \\
S2 & 70.8 & 65.7 & 43.1 & 59.8 & 41.3 & 40.6 & 63.5 & 65.7 & 42.3 & 70.4 \\
OLI & 66.2 & 63.4 & 42.2 & 53.3 & 40.1 & 39.6 & 55.8 & 63.5 & 42.6 & 67.3 \\
\midrule
All HSI & \underline{73.3} & \underline{70.9} & \textbf{45.0} & \textbf{63.7} & \underline{43.5} & \underline{42.0} & \underline{65.8} & \textbf{69.0} & \underline{43.0} & \underline{78.6} \\
\midrule
All HSI + S2 + OLI & \textbf{74.0} & \textbf{70.9} & \underline{44.9} & \underline{63.5} & \textbf{43.7} & \textbf{42.1} & \textbf{66.3} & 68.0 & \textbf{43.2} & \textbf{78.9} \\
\bottomrule
\end{tabular}
\end{table*}

\paragraph{Co-located multisensor fusion}

\Cref{tab:real-fusion} reports results with co-located acquisitions, which are not directly comparable to \Cref{tab:hsi-main} due to reduced scene coverage. On BD-Forêt, EnMAP outperforms S2 as a single sensor, consistent with tree species discrimination relying on fine spectral resolution. On CDL and EuroCrops, EnMAP and S2 score within $0.4$~mIoU of each other: S2's red-edge bands and higher spatial resolution compensate for its lower spectral dimensionality. OLI scores substantially lower on all three tasks, reflecting its coarser spectral sampling and the absence of red-edge channels. Fusing EnMAP with S2 improves on all three tasks, with the largest gain on EuroCrops ($+3.4$ mIoU over EnMAP). Adding OLI yields a further improvement across all tasks. %(CDL: $+0.1$, BD-Forêt: $+0.9$, EuroCrops: $+0.2$~mIoU)

\label{sec:real_fusion}

\begin{table}[t]
\centering
\caption{Sensor-fusion performance. Scores are mIoU (\%); best is \textbf{bold} and second best is \underline{underlined}.}
\label{tab:real-fusion}
\footnotesize
\setlength{\tabcolsep}{4pt}
\begin{tabular}{lccccccc}
\toprule
Dataset & EnMAP & S2 & OLI & EnMAP+S2 & EnMAP+OLI & S2+OLI & EnMAP+S2+OLI \\
\midrule
CDL & 70.8 & 70.7 & 66.1 & \underline{72.9} & 71.9 & 71.5 & \textbf{73.0} \\
BD-F & 71.1 & 69.4 & 64.4 & \underline{72.9} & 72.2 & 71.1 & \textbf{73.8} \\
EuroCrops & 66.8 & 66.4 & 59.3 & \underline{70.2} & 68.2 & 67.1 & \textbf{70.4} \\
\bottomrule
\end{tabular}
\end{table}

%% -------------------------------------------------------
\subsection{General EO benchmarks}
\label{sec:pangaea}

We evaluate on seven PANGAEA tasks \citep{marsocci2024pangaea}: HLS Burns \citep{jakubik2023prithvi}, MADOS \citep{kikaki2024detecting},
PASTIS-R \citep{garnot2022multi,garnot2021panoptic}, Sen1Floods11 \citep{rambour2020flood}, DynEarthNet \citep{toker2022dynamicearthnet}, SN7 \citep{van2018spacenet}, and AI4Farms \citep{parsello2023ai4smallfarms},
following the official protocol. We were unable to reproduce
results for three tasks (Crop Type Mapping-SS~\citep{m2019semantic},
BioMassters~\citep{nascetti2023biomassters} and FiveBillionPixels~\citep{tong2023enabling}) and therefore exclude them. Baseline scores are taken from \citep{marsocci2024pangaea,forgaard2026thor}.

\paragraph{Results}
\Cref{tab:pangaea} reports mIoU at $100\%$ training data on PANGAEA. \modelname-a achieves an average rank of $3.43$, the best in the comparison, while \modelname-s ranks $4.86$, close to Terramind. The multi-branch variant improves over the single-branch variant on most tasks, except for MADOS and PASTIS. MADOS contains fine-grained segmentation labels, and the lower spatial resolution of OLI ($30~m$ vs.\ $10~m$ for S2) degrades performance. Together with \Cref{sec:hsi_eval}, these results show that \modelname achieves competitive performance across both settings, ranking first on hyperspectral benchmarks and among the top models on standard EO benchmarks.

\begin{table}[th!]
\centering
\caption{PANGAEA results at 100\% training data. U-Net and ViT baselines are trained from scratch. Best and second-best per column are in \textbf{bold} and \underline{underlined}.}
\label{tab:pangaea}
\small
\setlength{\tabcolsep}{3pt}
\renewcommand{\arraystretch}{0.89}
\begin{adjustbox}{max width=\textwidth}
\begin{tabular}{@{}>{\raggedright\arraybackslash}p{0.26\textwidth}cccccccc@{}}
\toprule
Model & \multicolumn{1}{c}{HLS Burns} & \multicolumn{1}{c}{MADOS} & \multicolumn{1}{c}{PASTIS} & \multicolumn{1}{c}{Sen1Floods11} & \multicolumn{1}{c}{DynEarthNet} & \multicolumn{1}{c}{SN7} & \multicolumn{1}{c}{AI4Farms} & \multicolumn{1}{c}{Avg. Rank} \\
\midrule
CROMA & 82.42 & \underline{67.55} & 32.32 & 90.89 & 38.29 & 59.28 & 25.65 & 6.43 \\
DOFA & 80.63 & 59.58 & 30.02 & 89.37 & \underline{39.29} & 61.84 & 27.07 & 7.86 \\
GFM-Swin & 76.90 & 64.71 & 21.24 & 72.60 & 34.09 & 60.89 & 27.19 & 11.57 \\
Prithvi & 83.62 & 49.98 & 33.93 & 90.37 & 27.86 & 56.54 & 26.86 & 10.86 \\
RemoteCLIP & 76.59 & 60.00 & 18.23 & 74.26 & 31.78 & 57.76 & 25.12 & 14.43 \\
SatlasNet & 79.96 & 55.86 & 17.51 & 90.30 & 36.31 & 61.88 & 25.13 & 10.86 \\
Scale-MAE & 76.68 & 57.32 & 24.55 & 74.13 & 35.11 & \textbf{62.96} & 21.47 & 12.86 \\
SpectralGPT & 80.47 & 57.99 & 35.44 & 89.07 & 37.85 & 58.86 & 26.75 & 9.00 \\
S12-MoCo & 81.58 & 51.76 & 34.49 & 89.26 & 35.44 & 57.64 & 25.38 & 11.29 \\
S12-DINO & 81.72 & 49.37 & 36.18 & 88.61 & 34.81 & 56.47 & 25.62 & 12.29 \\
S12-MAE & 81.91 & 49.90 & 32.03 & 87.79 & 34.08 & 57.13 & 24.69 & 13.57 \\
S12-Data2Vec & 81.91 & 44.36 & 34.32 & 88.15 & 35.90 & 58.23 & 24.23 & 12.43 \\
TerraMindv1-B & 82.42 & \textbf{69.52} & 40.51 & 90.62 & 37.87 & 60.61 & 28.12 & \underline{4.29} \\
THOR-B & 79.65 & 51.48 & 40.76 & 89.44 & 37.57 & 59.87 & 26.29 & 9.29 \\
\midrule
U-Net Baseline & \underline{84.51} & 54.79 & 31.60 & \textbf{91.42} & \textbf{39.46} & 62.09 & \textbf{46.34} & 4.57 \\
ViT Baseline & 81.58 & 48.19 & 38.53 & 87.66 & 36.83 & 52.57 & \underline{38.37} & 10.71 \\
\midrule
\textbf{SpectralEarth-FM-s} & 84.35 & 64.02 & \textbf{44.85} & 90.44 & 36.00 & 61.79 & 27.53 & 4.86 \\
\textbf{SpectralEarth-FM-a} & \textbf{85.10} & 57.52 & \underline{43.28} & \underline{91.08} & 37.77 & \underline{62.16} & 28.44 & \textbf{3.43} \\
\bottomrule
\end{tabular}
\end{adjustbox}
\end{table}

\section{Conclusions \& future work}
\label{sec:conclusion}
We introduced \modelname, a multisensor transformer-based EO foundation model, and \datasetname, a large-scale dataset co-locating HSI from three spaceborne sensors with multispectral, SAR, and LST observations. \modelname combines spectral tokenization, sensor-specific branches, hierarchical spatial processing, and a shared encoder pretrained with a JEPA-style objective. Evaluated across ten hyperspectral benchmarks and the PANGAEA suite, \modelname achieves state-of-the-art results in both settings. Both \modelname and \datasetname will be released to support future research in EO foundation models.

\paragraph{Limitations}Unlike any-sensor models, \modelname dedicates explicit processing to each pretraining sensor, trading out-of-the-box sensor flexibility for stronger in-distribution spectral representations. As shown in \cref{tab:real-fusion} and \cref{sec:hsi_eval}, fusing sensors with varying spatio-spectral characteristics involves a complex interplay of spatial and spectral features that is not yet fully understood. Future work should investigate explainability methods to understand how these correlations are encoded in the shared representation, and extend the framework beyond land surface imagery to oceanic and atmospheric applications.

\paragraph{Code and data availability}
Code, pretrained checkpoints, and data will be made available.
The \datasetname data release will follow the licenses and access conditions
of the source missions.

{\small
\section*{Acknowledgement}
NAAB and CMA received funding from the European Commission through
the \textit{EvoLand} project under the Horizon~2020 Research and
Innovation Program (Grant Agreement No.~\texttt{101082130}) until
November~2025. Thereafter, \textit{Embed2Scale} funded the work of
NAAB and CMA through the EU Horizon Europe Program under Grant
Agreement No.~\texttt{101131841}. Additional funding for
\textit{Embed2Scale} was provided by the Swiss State Secretariat for
Education, Research and Innovation (SERI) and UK Research and
Innovation (UKRI). AB was supported by the HYPER-AMPLIFAI project
funded by the Helmholtz Association of German Research Centres (HGF),
contract number ZT-I-PF-4-056. CMA is grateful for Columbia
University's hospitality in 2025 and 2026. JM was supported by ERC
grant No.~101087696 (APHELEIA project). JC was supported by the MIAI
Cluster grant ANR-23-IACL-0006.
}

\clearpage
\bibliographystyle{abbrvnat}  % abbreviates first names
\bibliography{references}

@inproceedings{
    cong2022satmae,
    title={Sat{MAE}: Pre-training Transformers for Temporal and Multi-Spectral Satellite Imagery},
    author={Yezhen Cong and Samar Khanna and Chenlin Meng and Patrick Liu and Erik Rozi and Yutong He and Marshall Burke and David B. Lobell and Stefano Ermon},
    booktitle={Advances in Neural Information Processing Systems},
    year={2022},
    editor={Alice H. Oh and Alekh Agarwal and Danielle Belgrave and Kyunghyun Cho},
    url={https://openreview.net/forum?id=WBhqzpF6KYH}
}

@inproceedings{fuller2023croma,
 author = {Fuller, Anthony and Millard, Koreen and Green, James},
 booktitle = {Advances in Neural Information Processing Systems},
 editor = {A. Oh and T. Naumann and A. Globerson and K. Saenko and M. Hardt and S. Levine},
 pages = {5506--5538},
 publisher = {Curran Associates, Inc.},
 title = {CROMA: Remote Sensing Representations with Contrastive Radar-Optical Masked Autoencoders},
 url = {https://proceedings.neurips.cc/paper_files/paper/2023/file/11822e84689e631615199db3b75cd0e4-Paper-Conference.pdf},
 volume = {36},
 year = {2023}
}

@article{tushar2026hyperfm,
  title={HyperFM: An Efficient Hyperspectral Foundation Model with Spectral Grouping},
  author={Tushar, Zahid Hassan and Purushotham, Sanjay},
  journal={arXiv preprint arXiv:2604.21127},
  year={2026}
}

@inproceedings{ryali2023hiera,
author = {Ryali, Chaitanya and Hu, Yuan-Ting and Bolya, Daniel and Wei, Chen and Fan, Haoqi and Huang, Po-Yao and Aggarwal, Vaibhav and Chowdhury, Arkabandhu and Poursaeed, Omid and Hoffman, Judy and Malik, Jitendra and Li, Yanghao and Feichtenhofer, Christoph},
title = {Hiera: a hierarchical vision transformer without the bells-and-whistles},
year = {2023},
publisher = {JMLR.org},
booktitle = {Proceedings of the 40th International Conference on Machine Learning},
articleno = {1224},
numpages = {14},
location = {Honolulu, Hawaii, USA},
series = {ICML'23}
}

@article{assran2023ijepa,
  title={Self-Supervised Learning from Images with a Joint-Embedding Predictive Architecture},
  author={Assran, Mahmoud and Duval, Quentin and Misra, Ishan and Bojanowski, Piotr and Vincent, Pascal and Rabbat, Michael and LeCun, Yann and Ballas, Nicolas},
  journal={arXiv preprint arXiv:2301.08243},
  year={2023}
}

@article{balestriero2025lejepa,
  title={Lejepa: Provable and scalable self-supervised learning without the heuristics},
  author={Balestriero, Randall and LeCun, Yann},
  journal={arXiv preprint arXiv:2511.08544},
  year={2025}
}

@article{xiong2024dofa,
  title={Neural plasticity-inspired multimodal foundation model for earth observation},
  author={Xiong, Zhitong and Wang, Yi and Zhang, Fahong and Stewart, Adam J and Hanna, Jo{\"e}lle and Borth, Damian and Papoutsis, Ioannis and Saux, Bertrand Le and Camps-Valls, Gustau and Zhu, Xiao Xiang},
  journal={arXiv preprint arXiv:2403.15356},
  year={2024}
}

@inproceedings{waldmann2025panopticon,
    author={Waldmann, Leonard and Shah, Ando and Wang, Yi and Lehmann, Nils and Stewart, Adam and Xiong, Zhitong and Zhu, Xiao Xiang and Bauer, Stefan and Chuang, John},
    title={Panopticon: Advancing Any-Sensor Foundation Models for Earth Observation},
    booktitle={Proceedings of the Computer Vision and Pattern Recognition Conference (CVPR) Workshops},
    year={2025},
    pages={2204-2214}
}

@inproceedings{
baumann2025carl,
title={{CARL}: Camera-Agnostic Representation Learning for Spectral Image Analysis},
author={Alexander Baumann and Leonardo Ayala and Silvia Seidlitz and Jan Sellner and Alexander Studier-Fischer and Berkin {\"O}zdemir and Lena Maier-hein and Slobodan Ilic},
booktitle={The Fourteenth International Conference on Learning Representations},
year={2026},
url={https://openreview.net/forum?id=TpbhS1yfz0}
}

@inproceedings{wang2025copernicusfm,
  author={Wang, Yi and Xiong, Zhitong and Liu, Chenying and Stewart, Adam J. and Dujardin, Thomas and Bountos, Nikolaos Ioannis and Zavras, Angelos and Gerken, Franziska and Papoutsis, Ioannis and Leal-Taixé, Laura and Zhu, Xiao Xiang},
  booktitle={2025 IEEE/CVF International Conference on Computer Vision (ICCV)}, 
  title={Towards a Unified Copernicus Foundation Model for Earth Vision}, 
  year={2025},
  volume={},
  number={},
  pages={9888-9899},
  doi={10.1109/ICCV51701.2025.00922}
  }

@inproceedings{astruc2025anysat,
  title={Anysat: One earth observation model for many resolutions, scales, and modalities},
  author={Astruc, Guillaume and Gonthier, Nicolas and Mallet, Clement and Landrieu, Loic},
  booktitle={Proceedings of the Computer Vision and Pattern Recognition Conference},
  pages={19530--19540},
  year={2025}
}

@inproceedings{jakubik2025terramind,
  author={Jakubik, Johannes and Yang, Felix and Blumenstiel, Benedikt and Scheurer, Erik and Sedona, Rocco and Maurogiovanni, Stefano and Bosmans, Jente and Dionelis, Nikolaos and Marsocci, Valerio and Kopp, Niklas and Ramachandran, Rahul and Fraccaro, Paolo and Brunschwiler, Thomas and Cavallaro, Gabriele and Bernabe-Moreno, Juan and Longépé, Nicolas},
  booktitle={IEEE/CVF International Conference on Computer Vision (ICCV)}, 
  title={TerraMind: Large-Scale Generative Multimodality for Earth Observation}, 
  year={2025},
  volume={},
  number={},
  pages={7383-7394},
  doi={10.1109/ICCV51701.2025.00693}
  }

@article{marsocci2024pangaea,
  author={Marsocci, Valerio and Jia, Yuru and Bellier, Georges Le and Kerekes, David and Zeng, Liang and Hafner, Sebastian and Gerard, Sebastian and Brune, Eric and Yadav, Ritu and Shibli, Ali and Fang, Heng and Ban, Yifang and Vergauwen, Maarten and Audebert, Nicolas and Nascetti, Andrea},
  journal={IEEE Geoscience and Remote Sensing Magazine}, 
  title={PANGAEA: Assessing Geospatial Foundation Models Capabilities through a Global and Inclusive Benchmark}, 
  year={2026},
  volume={14},
  number={1},
  pages={245-285},
  doi={10.1109/MGRS.2025.3628194}
  }

@article{jakubik2023prithvi,
  title={Foundation models for generalist geospatial artificial intelligence},
  author={Jakubik, Johannes and Roy, Sujit and Phillips, CE and Fraccaro, Paolo and Godwin, Denys and Zadrozny, Bianca and Szwarcman, Daniela and Gomes, Carlos and Nyirjesy, Gabby and Edwards, Blair and others},
  journal={arXiv preprint arXiv:2310.18660},
  year={2023}
}

@article{ji2025specaware,
title = {SpecAware: A Spectral-Content Aware Foundation Model for Unifying Multi-Sensor Learning in Hyperspectral Remote Sensing Mapping},
journal = {ISPRS Journal of Photogrammetry and Remote Sensing},
year = {2026},
author = {Renjie Ji and Xue Wang and Chao Niu and Wen Zhang and Yong Mei and Kun Tan},
volume = {234},
pages = {242-260},
issn = {0924-2716},
doi = {https://doi.org/10.1016/j.isprsjprs.2026.02.024},
url = {https://www.sciencedirect.com/science/article/pii/S0924271626000754},
}

@inproceedings{sumbul2025smarties,
  title={SMARTIES: Spectrum-aware multi-sensor auto-encoder for remote sensing images},
  author={Sumbul, Gencer and Xu, Chang and Dalsasso, Emanuele and Tuia, Devis},
  booktitle={Proceedings of the IEEE/CVF International Conference on Computer Vision},
  pages={5569--5578},
  year={2025}
}

@article{szwarcman2024prithvi2,
  title={Prithvi-EO-2.0: A Versatile Multitemporal Foundation Model for Earth Observation Applications},
  author={Szwarcman, Daniela and Roy, Sujit and Fraccaro, Paolo and G{\'\i}slason, Orsteinn El{\'\i} and Blumenstiel, Benedikt and Ghosal, Rinki and De Oliveira, Pedro Henrique and de Sousa Almeida, Joao Lucas and Sedona, Rocco and Kang, Yanghui and others},
  journal={IEEE Transactions on Geoscience and Remote Sensing},
  year={2025},
  publisher={IEEE},
  doi={10.1109/TGRS.2025.3642610},
  volume={64},
  number={},
  pages={1-20},
}

@article{wang2024hypersigma,
  author={Wang, Di and Hu, Meiqi and Jin, Yao and Miao, Yuchun and Yang, Jiaqi and Xu, Yichu and Qin, Xiaolei and Ma, Jiaqi and Sun, Lingyu and Li, Chenxing and Fu, Chuan and Chen, Hongruixuan and Han, Chengxi and Yokoya, Naoto and Zhang, Jing and Xu, Minqiang and Liu, Lin and Zhang, Lefei and Wu, Chen and Du, Bo and Tao, Dacheng and Zhang, Liangpei},
  journal={IEEE Transactions on Pattern Analysis and Machine Intelligence}, 
  title={HyperSIGMA: Hyperspectral Intelligence Comprehension Foundation Model}, 
  year={2025},
  volume={47},
  number={8},
  pages={6427-6444},
  doi={10.1109/TPAMI.2025.3557581}
}

@article{kong2025hypersl,
  author={Kong, Weili and Liu, Baisen and Bi, Xiaojun and Yu, Changdong and Li, Xinyao and Chen, Yushi},
  journal={IEEE Transactions on Geoscience and Remote Sensing}, 
  title={HyperSL: A Spectral Foundation Model for Hyperspectral Image Interpretation}, 
  year={2025},
  volume={63},
  number={},
  pages={1-19},
  doi={10.1109/TGRS.2025.3566205}
}

@article{braham2024spectralearth,
  author={Braham, Nassim Ait Ali and Albrecht, Conrad M. and Mairal, Julien and Chanussot, Jocelyn and Wang, Yi and Zhu, Xiao Xiang},
  journal={IEEE Journal of Selected Topics in Applied Earth Observations and Remote Sensing}, 
  title={SpectralEarth: Training Hyperspectral Foundation Models at Scale}, 
  year={2025},
  volume={18},
  number={},
  pages={16780-16797},
  doi={10.1109/JSTARS.2025.3581451}
}

@inproceedings{chen2021exploring,
  author={Chen, Xinlei and He, Kaiming},
  booktitle={IEEE/CVF Conference on Computer Vision and Pattern Recognition (CVPR)}, 
  title={Exploring Simple Siamese Representation Learning}, 
  year={2021},
  volume={},
  number={},
  pages={15745-15753},
  doi={10.1109/CVPR46437.2021.01549}
}

@article{grill2020bootstrap,
  title={Bootstrap your own latent-a new approach to self-supervised learning},
  author={Grill, Jean-Bastien and Strub, Florian and Altch{\'e}, Florent and Tallec, Corentin and Richemond, Pierre and Buchatskaya, Elena and Doersch, Carl and Avila Pires, Bernardo and Guo, Zhaohan and Gheshlaghi Azar, Mohammad and others},
  journal={Advances in neural information processing systems},
  volume={33},
  pages={21271--21284},
  year={2020}
}

@inproceedings{he2022masked,
  title={Masked autoencoders are scalable vision learners},
  author={He, Kaiming and Chen, Xinlei and Xie, Saining and Li, Yanghao and Doll{\'a}r, Piotr and Girshick, Ross},
  booktitle={Proceedings of the IEEE/CVF conference on computer vision and pattern recognition},
  pages={16000--16009},
  year={2022}
}

@article{hong2022spectralformer,
  author={Hong, Danfeng and Han, Zhu and Yao, Jing and Gao, Lianru and Zhang, Bing and Plaza, Antonio and Chanussot, Jocelyn},
  journal={IEEE Transactions on Geoscience and Remote Sensing}, 
  title={SpectralFormer: Rethinking Hyperspectral Image Classification With Transformers}, 
  year={2022},
  volume={60},
  number={},
  pages={1-15},
  doi={10.1109/TGRS.2021.3130716}
}

@article{wang2022self,
  author={Wang, Yi and Albrecht, Conrad M. and Braham, Nassim Ait Ali and Mou, Lichao and Zhu, Xiao Xiang},
  journal={IEEE Geoscience and Remote Sensing Magazine}, 
  title={Self-Supervised Learning in Remote Sensing: A review}, 
  year={2022},
  volume={10},
  number={4},
  pages={213-247},
  doi={10.1109/MGRS.2022.3198244}
}

@article{pearlman2003hyperion,
  author={Pearlman, J.S. and Barry, P.S. and Segal, C.C. and Shepanski, J. and Beiso, D. and Carman, S.L.},
  journal={IEEE Transactions on Geoscience and Remote Sensing}, 
  title={Hyperion, a space-based imaging spectrometer}, 
  year={2003},
  volume={41},
  number={6},
  pages={1160-1173},
  doi={10.1109/TGRS.2003.815018}
}

@article{chen2023spectral,
  author={Chen, Huan and Zhao, Wangcai and Xu, Tingfa and Shi, Guokai and Zhou, Shiyun and Liu, Peifu and Li, Jianan},
  journal={IEEE Transactions on Circuits and Systems for Video Technology}, 
  title={Spectral-Wise Implicit Neural Representation for Hyperspectral Image Reconstruction}, 
  year={2024},
  volume={34},
  number={5},
  pages={3714-3727},
  doi={10.1109/TCSVT.2023.3318366}
}

@article{sun2022ringmo,
  author={Sun, Xian and Wang, Peijin and Lu, Wanxuan and Zhu, Zicong and Lu, Xiaonan and He, Qibin and Li, Junxi and Rong, Xuee and Yang, Zhujun and Chang, Hao and He, Qinglin and Yang, Guang and Wang, Ruiping and Lu, Jiwen and Fu, Kun},
  journal={IEEE Transactions on Geoscience and Remote Sensing}, 
  title={RingMo: A Remote Sensing Foundation Model With Masked Image Modeling}, 
  year={2023},
  volume={61},
  number={},
  pages={1-22},
  doi={10.1109/TGRS.2022.3194732}
}

@article{yao2023ringmo,
  author={Yao, Fanglong and Lu, Wanxuan and Yang, Heming and Xu, Liangyu and Liu, Chenglong and Hu, Leiyi and Yu, Hongfeng and Liu, Nayu and Deng, Chubo and Tang, Deke and Chen, Changshuo and Yu, Jiaqi and Sun, Xian and Fu, Kun},
  journal={IEEE Transactions on Geoscience and Remote Sensing}, 
  title={RingMo-Sense: Remote Sensing Foundation Model for Spatiotemporal Prediction via Spatiotemporal Evolution Disentangling}, 
  year={2023},
  volume={61},
  number={},
  pages={1-21},
  doi={10.1109/TGRS.2023.3316166}
}

@article{liu2019advanced,
  author={Liu, Yin-Nian and Sun, De-Xin and Hu, Xiao-Ning and Ye, Xiang and Li, Yun-Duan and Liu, Shu-Feng and Cao, Kai-Qin and Chai, Meng-Yang and Zhou, Wei-Yi-Nuo and Zhang, Jing and Zhang, Ying and Sun, Wei-Wei and Jiao, Lei-Lei},
  journal={IEEE Geoscience and Remote Sensing Magazine}, 
  title={The Advanced Hyperspectral Imager: Aboard China's GaoFen-5 Satellite}, 
  year={2019},
  volume={7},
  number={4},
  pages={23-32},
  doi={10.1109/MGRS.2019.2927687}
}

@inproceedings{bastani2023satlaspretrain,
  author={Bastani, Favyen and Wolters, Piper and Gupta, Ritwik and Ferdinando, Joe and Kembhavi, Aniruddha},
  booktitle={IEEE/CVF International Conference on Computer Vision (ICCV)}, 
  title={SatlasPretrain: A Large-Scale Dataset for Remote Sensing Image Understanding}, 
  year={2023},
  volume={},
  number={},
  pages={16726-16736},
  doi={10.1109/ICCV51070.2023.01538}
}

@inproceedings{francis2024majortom,
  author={Francis, Alistair and Czerkawski, Mikolaj},
  booktitle={2024 IEEE International Geoscience and Remote Sensing Symposium}, 
  title={Major TOM: Expandable Datasets for Earth Observation}, 
  year={2024},
  volume={},
  number={},
  pages={2935-2940},
  doi={10.1109/IGARSS53475.2024.10640760}
}

@inproceedings{nedungadi2024mmearth,
  title={MMEarth: Exploring multi-modal pretext tasks for geospatial representation learning},
  author={Nedungadi, Vishal and Kariryaa, Ankit and Oehmcke, Stefan and Belongie, Serge and Igel, Christian and Lang, Nico},
  booktitle={European Conference on Computer Vision},
  pages={164--182},
  year={2024},
  organization={Springer}
}

@inproceedings{ayush2021iccv-geographyaware,
  author={Ayush, Kumar and Uzkent, Burak and Meng, Chenlin and Tanmay, Kumar and Burke, Marshall and Lobell, David and Ermon, Stefano},
  booktitle={IEEE/CVF International Conference on Computer Vision (ICCV)}, 
  title={Geography-Aware Self-Supervised Learning}, 
  year={2021},
  volume={},
  number={},
  pages={10161-10170},
  doi={10.1109/ICCV48922.2021.01002}
}

@inproceedings{manas2021seasonalcontrast,
  title={Seasonal contrast: Unsupervised pre-training from uncurated remote sensing data},
  author={Manas, Oscar and Lacoste, Alexandre and Gir{\'o}-i-Nieto, Xavier and Vazquez, David and Rodriguez, Pau},
  booktitle={Proceedings of the IEEE/CVF international conference on computer vision},
  pages={9414--9423},
  year={2021}
}

@inproceedings{danish2025terrafm,
title={Terra{FM}: A Scalable Foundation Model for Unified Multisensor Earth Observation},
author={Muhammad Sohail Danish and Muhammad Akhtar Munir and Syed Roshaan Ali Shah and Muhammad Haris Khan and Rao Muhammad Anwer and Jorma Laaksonen and Fahad Shahbaz Khan and Salman Khan},
booktitle={The Fourteenth International Conference on Learning Representations},
year={2026},
}

@article{blumenstiel2025terramesh,
  title={Terramesh: A planetary mosaic of multimodal earth observation data},
  author={Blumenstiel, Benedikt and Fraccaro, Paolo and Marsocci, Valerio and Jakubik, Johannes and Maurogiovanni, Stefano and Czerkawski, Mikolaj and Sedona, Rocco and Cavallaro, Gabriele and Brunschwiler, Thomas and Bernabe-Moreno, Juan and others},
  journal={Proceedings of the IEEE/CVF Conference on Computer Vision and Pattern Recognition (CVPR) Workshops},
  year={2025},
}

@inproceedings{fuchs2023hyspecnet,
  author={Fuchs, Martin Hermann Paul and Demir, Begüm},
  booktitle={2023 IEEE International Geoscience and Remote Sensing Symposium}, 
  title={HySpecNet-11k: a Large-Scale Hyperspectral Dataset for Benchmarking Learning-Based Hyperspectral Image Compression Methods}, 
  year={2023},
  volume={},
  number={},
  pages={1779-1782},
  doi={10.1109/IGARSS52108.2023.10283385}
}

@article{brown2025alphaearth,
  title={Alphaearth foundations: An embedding field model for accurate and efficient global mapping from sparse label data},
  author={Brown, Christopher F and Kazmierski, Michal R and Pasquarella, Valerie J and Rucklidge, William J and Samsikova, Masha and Zhang, Chenhui and Shelhamer, Evan and Lahera, Estefania and Wiles, Olivia and Ilyushchenko, Simon and others},
  journal={arXiv preprint arXiv:2507.22291},
  year={2025}
}

@inproceedings{astruc2024omnisat,
  title={Omnisat: Self-supervised modality fusion for earth observation},
  author={Astruc, Guillaume and Gonthier, Nicolas and Mallet, Clement and Landrieu, Loic},
  booktitle={European Conference on Computer Vision},
  pages={409--427},
  year={2024},
  organization={Springer}
}

@inproceedings{scheibenreif2023cvprw-masked,
  author={Scheibenreif, Linus and Mommert, Michael and Borth, Damian},
  booktitle={2023 IEEE/CVF Conference on Computer Vision and Pattern Recognition Workshops (CVPRW)}, 
  title={Masked Vision Transformers for Hyperspectral Image Classification}, 
  year={2023},
  volume={},
  number={},
  pages={2166-2176},
  doi={10.1109/CVPRW59228.2023.00210}
}

@ARTICLE{hong2024spectralgpt,
  author={Hong, Danfeng and Zhang, Bing and Li, Xuyang and Li, Yuxuan and Li, Chenyu and Yao, Jing and Yokoya, Naoto and Li, Hao and Ghamisi, Pedram and Jia, Xiuping and Plaza, Antonio and Gamba, Paolo and Benediktsson, Jon Atli and Chanussot, Jocelyn},
  journal={IEEE Transactions on Pattern Analysis and Machine Intelligence}, 
  title={SpectralGPT: Spectral Remote Sensing Foundation Model}, 
  year={2024},
  volume={46},
  number={8},
  pages={5227-5244},
  doi={10.1109/TPAMI.2024.3362475}
}

@inproceedings{xie2021segformer,
  title={SegFormer: Simple and Efficient Design for Semantic Segmentation with Transformers},
  author={Xie, Enze and Wang, Wenhai and Yu, Zhiding and Anandkumar, Anima and Alvarez, Jose M and Luo, Ping},
  booktitle={Advances in Neural Information Processing Systems},
  year={2021},
  volume={34},
  pages={12077--12090},
}

@article{forgaard2026thor,
  title={THOR: A Versatile Foundation Model for Earth Observation Climate and Society Applications},
  author={Forgaard, Theodor and Reksten, Jarle H and Waldeland, Anders U and Marsocci, Valerio and Long{\'e}p{\'e}, Nicolas and Kampffmeyer, Michael and Salberg, Arnt-B{\o}rre},
  journal={arXiv preprint arXiv:2601.16011},
  year={2026}
}

@inproceedings{m2019semantic,
  title={Semantic segmentation of crop type in Africa: A novel dataset and analysis of deep learning methods},
  author={M Rustowicz, Rose and Cheong, Robin and Wang, Lijing and Ermon, Stefano and Burke, Marshall and Lobell, David},
  booktitle={Proceedings of the IEEE/cvf conference on computer vision and pattern recognition workshops},
  pages={75--82},
  year={2019}
}

@article{nascetti2023biomassters,
  title={Biomassters: A benchmark dataset for forest biomass estimation using multi-modal satellite time-series},
  author={Nascetti, Andrea and Yadav, Ritu and Brodt, Kirill and Qu, Qixun and Fan, Hongwei and Shendryk, Yuri and Shah, Isha and Chung, Christine},
  journal={Advances in Neural Information Processing Systems},
  volume={36},
  pages={20409--20420},
  year={2023}
}

@article{wang2023ssl4eo,
  author={Wang, Yi and Braham, Nassim Ait Ali and Xiong, Zhitong and Liu, Chenying and Albrecht, Conrad M. and Zhu, Xiao Xiang},
  journal={IEEE Geoscience and Remote Sensing Magazine}, 
  title={SSL4EO-S12: A large-scale multimodal, multitemporal dataset for self-supervised learning in Earth observation [Software and Data Sets]}, 
  year={2023},
  volume={11},
  number={3},
  pages={98-106},
  doi={10.1109/MGRS.2023.3281651}
}

@article{tong2023enabling,
  title={Enabling country-scale land cover mapping with meter-resolution satellite imagery},
  author={Tong, Xin-Yi and Xia, Gui-Song and Zhu, Xiao Xiang},
  journal={ISPRS Journal of Photogrammetry and Remote Sensing},
  volume={196},
  pages={178--196},
  year={2023},
  publisher={Elsevier}
}

@article{kikaki2024detecting,
  title={Detecting marine pollutants and sea surface features with deep learning in sentinel-2 imagery},
  author={Kikaki, Katerina and Kakogeorgiou, Ioannis and Hoteit, Ibrahim and Karantzalos, Konstantinos},
  journal={ISPRS Journal of Photogrammetry and Remote Sensing},
  volume={210},
  pages={39--54},
  year={2024},
  publisher={Elsevier}
}

@inproceedings{garnot2021panoptic,
  title={Panoptic segmentation of satellite image time series with convolutional temporal attention networks},
  author={Garnot, Vivien Sainte Fare and Landrieu, Loic},
  booktitle={Proceedings of the IEEE/CVF International Conference on Computer Vision},
  pages={4872--4881},
  year={2021}
}

@article{garnot2022multi,
  title={Multi-modal temporal attention models for crop mapping from satellite time series},
  author={Garnot, Vivien Sainte Fare and Landrieu, Loic and Chehata, Nesrine},
  journal={ISPRS Journal of Photogrammetry and Remote Sensing},
  volume={187},
  pages={294--305},
  year={2022},
  publisher={Elsevier}
}

@article{rambour2020flood,
  title={Flood detection in time series of optical and sar images},
  author={Rambour, Cl{\'e}ment and Audebert, Nicolas and Koeniguer, E and Le Saux, Bertrand and Crucianu, M and Datcu, Mihai},
  journal={The International Archives of the Photogrammetry, Remote Sensing and Spatial Information Sciences},
  volume={43},
  number={B2},
  pages={1343--1346},
  year={2020}
}

@inproceedings{toker2022dynamicearthnet,
  title={Dynamicearthnet: Daily multi-spectral satellite dataset for semantic change segmentation},
  author={Toker, Aysim and Kondmann, Lukas and Weber, Mark and Eisenberger, Marvin and Camero, Andr{\'e}s and Hu, Jingliang and Hoderlein, Ariadna Pregel and {\c{S}}enaras, {\c{C}}a{\u{g}}lar and Davis, Timothy and Cremers, Daniel and others},
  booktitle={Proceedings of the IEEE/CVF Conference on Computer Vision and Pattern Recognition},
  pages={21158--21167},
  year={2022}
}

@article{voautomatic,
  title={Automatic Data Curation for Self-Supervised Learning: A Clustering-Based Approach},
  author={Vo, Huy V and Khalidov, Vasil and Darcet, Timoth{\'e}e and Moutakanni, Th{\'e}o and Smetanin, Nikita and Szafraniec, Marc and Touvron, Hugo and Oquab, Maxime and Joulin, Armand and Jegou, Herve and others},
  journal={Transactions on Machine Learning Research},
  year=2024,
}

@article{van2018spacenet,
  title={Spacenet: A remote sensing dataset and challenge series},
  author={Van Etten, Adam and Lindenbaum, Dave and Bacastow, Todd M},
  journal={arXiv preprint arXiv:1807.01232},
  year={2018}
}

@ARTICLE{parsello2023ai4smallfarms,
  author={Persello, Claudio and Grift, Jeroen and Fan, Xinyan and Paris, Claudia and Hänsch, Ronny and Koeva, Mila and Nelson, Andrew},
  journal={IEEE Geoscience and Remote Sensing Letters}, 
  title={AI4SmallFarms: A Dataset for Crop Field Delineation in Southeast Asian Smallholder Farms}, 
  year={2023},
  volume={20},
  number={},
  pages={1-5},
  doi={10.1109/LGRS.2023.3323095}}

@ARTICLE{ruzicka2025hyperspectralvits,
  author={Růžička, Vít and Markham, Andrew},
  journal={IEEE Journal of Selected Topics in Applied Earth Observations and Remote Sensing}, 
  title={HyperspectralViTs: General Hyperspectral Models for On-Board Remote Sensing}, 
  year={2025},
  volume={18},
  number={},
  pages={10241-10253},
  doi={10.1109/JSTARS.2025.3557527}}

@article{guanter2015enmap,
AUTHOR = {Guanter, Luis and Kaufmann, Hermann and Segl, Karl and Foerster, Saskia and Rogass, Christian and Chabrillat, Sabine and Kuester, Theres and Hollstein, André and Rossner, Godela and Chlebek, Christian and Straif, Christoph and Fischer, Sebastian and Schrader, Stefanie and Storch, Tobias and Heiden, Uta and Mueller, Andreas and Bachmann, Martin and Mühle, Helmut and Müller, Rupert and Habermeyer, Martin and Ohndorf, Andreas and Hill, Joachim and Buddenbaum, Henning and Hostert, Patrick and Van der Linden, Sebastian and Leitão, Pedro J. and Rabe, Andreas and Doerffer, Roland and Krasemann, Hajo and Xi, Hongyan and Mauser, Wolfram and Hank, Tobias and Locherer, Matthias and Rast, Michael and Staenz, Karl and Sang, Bernhard},
TITLE = {The EnMAP Spaceborne Imaging Spectroscopy Mission for Earth Observation},
JOURNAL = {Remote Sensing},
VOLUME = {7},
YEAR = {2015},
NUMBER = {7},
PAGES = {8830--8857},
URL = {https://www.mdpi.com/2072-4292/7/7/8830},
ISSN = {2072-4292},
DOI = {10.3390/rs70708830}
}

@article{hong2023cross,
  title={Cross-city matters: A multimodal remote sensing benchmark dataset for cross-city semantic segmentation using high-resolution domain adaptation networks},
  author={Hong, Danfeng and Zhang, Bing and Li, Hao and Li, Yuxuan and Yao, Jing and Li, Chenyu and Werner, Martin and Chanussot, Jocelyn and Zipf, Alexander and Zhu, Xiao Xiang},
  journal={Remote Sensing of Environment},
  volume={299},
  pages={113856},
  year={2023},
  publisher={Elsevier}
}

@misc{nasa_esd_dis_policy,
  author       = {{NASA Earth Science Data and Information System}},
  title        = {Data Use and Citation Guidance for Earth Science Data},
  year         = {2025},
  url          = {https://doi.org/10.5067/DOC/ESCO/ESDS-RFC-055},
  note         = {NASA Earth Science data are fully open access without use restrictions, following the ESDS-RFC-055 standard.}
}

@misc{desis_license,
  author       = {{German Aerospace Center (DLR)}},
  title        = {License Agreement Regarding the Use of the {DESIS} Data for Scientific Use},
  year         = {2024},
  url          = {https://geoservice.dlr.de/resources/licenses/desis/DESIS_License_Agreement_for_Scientific_Use.pdf},
  note         = {Free for non-commercial scientific research; commercial use managed by Teledyne Brown Engineering.}
}

@misc{enmap_geoservice,
  author       = {{German Aerospace Center (DLR)}},
  title        = {{EnMAP} - Environmental Mapping and Analysis Program Data Policy and Access},
  year         = {2023},
  url          = {https://www.enmap.org/data_access/},
  howpublished = {\url{https://www.enmap.org/data/resources/EnMAP_Data_License.pdf}},
  note         = {Scientific and commercial use permitted as per the EnMAP Data License Agreement.}
}

@misc{copernicus_data_policy,
  author       = {{European Union}},
  title        = {Copernicus Legal Notice: Free, Full and Open Access to {Sentinel} Data},
  year         = {2024},
  url          = {https://www.copernicus.eu/en/terms-use/how-access-data},
  note         = {Covers Sentinel-1 and Sentinel-2 data access and exploitation for any public or private organization.}
}

@misc{usgs_landsat_data_policy,
 author = {{U.S. Geological Survey}},
  title  = {Are Landsat data in the cloud still considered to be within the public domain?},
  year   = {2020},
  url    = {https://www.usgs.gov/faqs/are-landsat-data-cloud-still-considered-be-within-public-domain},
  note   = {Accessed: 2026-05-20}
}

@article{Bohnetal2024,
  title={Full mission evaluation of EnMAP water leaving reflectance products using three atmospheric correction processors},
  author={Soppa, Mariana A and Brell, Maximilian and Chabrillat, Sabine and Alvarado, Leonardo MA and Gege, Peter and Plattner, Stefan and Somlai-Schweiger, Ian and Schroeder, Thomas and Steinmetz, Fran{\c{c}}ois and Scheffler, Daniel and others},
  journal={Optics Express},
  volume={32},
  number={16},
  pages={28215--28230},
  year={2024},
  publisher={Optica Publishing Group}
}

@manual{DLR_EnMAP_FAQ,
  title        = {{EnMAP} Frequently Asked Questions ({FAQ})},
  author       = {{German Aerospace Center (DLR)}},
  organization = {German Remote Sensing Data Center},
  year         = {2026},
  edition      = {2.7},
  url          = {https://www.enmap.org/data/doc/EnMAP_FAQ.pdf},
}

@misc{NASA_EMIT_L2A_Guide,
  author       = {{NASA LP DAAC}},
  title        = {{EMIT} {L2A} Estimated Surface Reflectance and Uncertainty and Masks 60 m {V001}},
  year         = {2025},
  howpublished = {NASA Earthdata Search},
  doi          = {10.5067/EMIT/EMITL2ARFL.001},
}

@article{Alonsoetal2019,
  title={Data products, quality and validation of the DLR earth sensing imaging spectrometer (DESIS)},
  author={Alonso, Kevin and Bachmann, Martin and Burch, Kara and Carmona, Emiliano and Cerra, Daniele and De los Reyes, Raquel and Dietrich, Daniele and Heiden, Uta and H{\"o}lderlin, Andreas and Ickes, Jack and others},
  journal={Sensors},
  volume={19},
  number={20},
  pages={4471},
  year={2019},
  publisher={MDPI}
}

@article{krutz2019instrument,
  title={The instrument design of the DLR earth sensing imaging spectrometer (DESIS)},
  author={Krutz, David and M{\"u}ller, Rupert and Knodt, Uwe and G{\"u}nther, Burghardt and Walter, Ingo and Sebastian, Ilse and S{\"a}uberlich, Thomas and Reulke, Ralf and Carmona, Emiliano and Eckardt, Andreas and others},
  journal={Sensors},
  volume={19},
  number={7},
  pages={1622},
  year={2019},
  publisher={MDPI}
}

@inproceedings{green2020earth,
  title={The Earth surface mineral dust source investigation: An Earth science imaging spectroscopy mission},
  author={Green, Robert O and Mahowald, Natalie and Ung, Charlene and Thompson, David R and Bator, Lori and Bennet, Matthew and Bernas, Michael and Blackway, Natalie and Bradley, Christine and Cha, Jeff and others},
  booktitle={2020 IEEE aerospace conference},
  pages={1--15},
  year={2020},
  organization={IEEE}
}

@article{leonardi2026spectral,
  title={Spectral Gaps and Spatial Priors: Studying Hyperspectral Downstream Adaptation Using TerraMind},
  author={Leonardi, Julia Anna and Jakubik, Johannes and Fraccaro, Paolo and Brovelli, Maria Antonia},
  journal={arXiv preprint arXiv:2603.06690},
  year={2026}
}

@misc{eoweb_geoportal,
  author       = {{German Aerospace Center (DLR)}},
  title        = {{EOWEB GeoPortal}},
  howpublished = {\url{https://eoweb.dlr.de/egp/}},
  year         = {2024},
  note         = {Accessed: 2025}
}

\newpage

\appendix
\section{Dataset details}
\label{app:dataset_details}

This appendix provides additional details on the construction of \datasetname. 

\subsection{Spectral coverage}

\begin{figure}[h!]
  \centering
  \includegraphics[width=\linewidth]{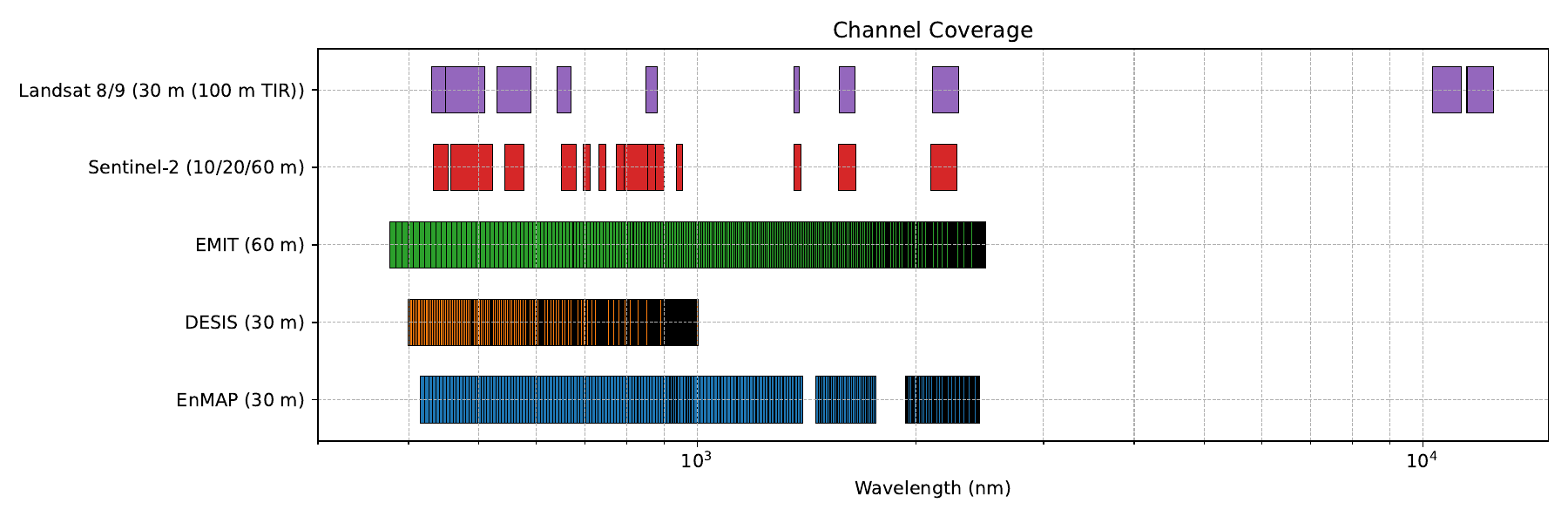}
  \caption{
  Spectral coverage of the optical sensors and Landsat thermal bands (long wavelength to the far right) associated with the LST product in \datasetname. \EnMAP, \EMIT, and \DESIS provide dense VNIR/SWIR measurements, while \Stwo and \Landsat provide sparse multispectral bands. \Sone is omitted because it is non-optical.
  }
  \label{fig:spectral_coverage}
\end{figure}

\cref{fig:spectral_coverage} illustrates the varying number of spectral bands, spectral coverage and spectral resolution of the optical sensors included in \datasetname.

\subsection{Construction pipeline}

\begin{figure}[t]
  \centering
  \includegraphics[width=1.0\linewidth]{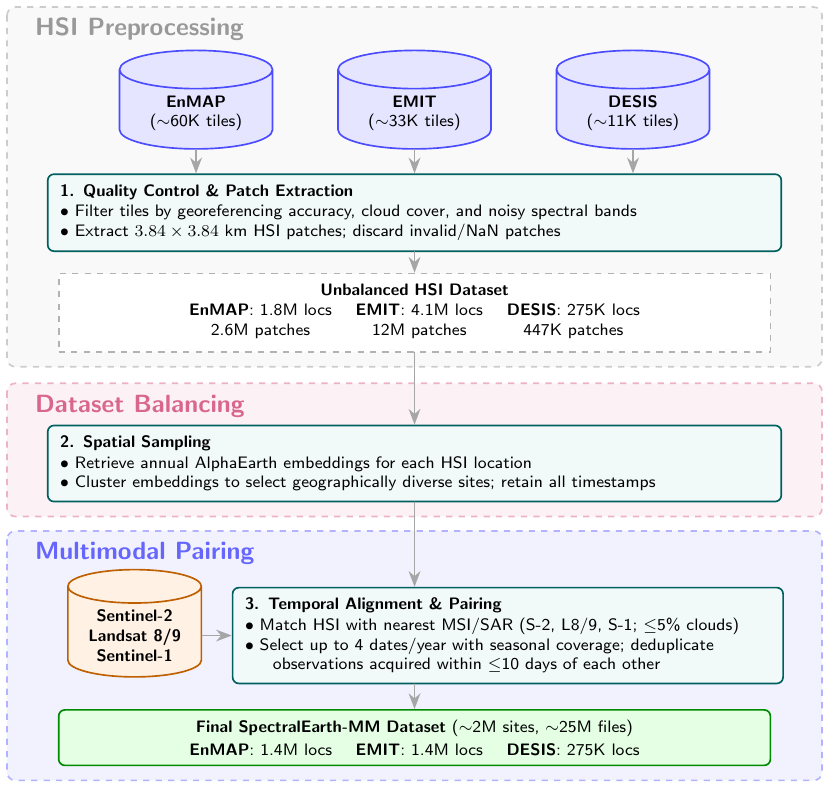}
  \caption{
  Construction pipeline for \datasetname. HSI acquisitions are used as anchors, filtered, patchified, grouped by location, rebalanced, and paired with co-located MSI, SAR, and LST observations. 
  }
  \label{fig:dataset_pipeline}
\end{figure}

\cref{fig:dataset_pipeline} provides an overview of the construction pipeline of \datasetname, which proceeds through the following stages: HSI preprocessing, dataset balancing and multimodal pairing.

\paragraph{HSI acquisition and quality control}
We assemble the HSI archive from \EnMAP, \EMIT, and \DESIS mission data. For \EnMAP, we use Level-2A hyperspectral products from 2022 to September 2025 and apply an initial 10\% cloud filter. For \DESIS, we use cloud-free tiles selected from the available archive via the EOWEB GeoPortal \citep{eoweb_geoportal}. For \EMIT, we collect available Level-2A imagery up to September 2025 with cloud coverage below 10\%. The resulting raw HSI archive contains approximately 60K \EnMAP tiles, 33K \EMIT tiles, and 11K \DESIS tiles.

Before patch extraction, we apply sensor-specific quality control. For \EnMAP and \DESIS, we filter scenes using the orthorectification RMSE reported in the metadata, with a threshold of 60 m. For \EnMAP and \EMIT, we additionally remove scenes with heavy cloud contamination not captured by the provided cloud metadata based on visual inspection. For all three HSI sensors, we remove noisy spectral bands, including bands affected by water absorption. \EMIT scenes are reprojected to local UTM zones to match the spatial reference convention used for \EnMAP and \DESIS.

\paragraph{Patch extraction and grouping}
After quality control, HSI scenes are patchified into non-overlapping $3.84\,\mathrm{km} \times 3.84\,\mathrm{km}$ footprints. Patches with more than 1\% missing or invalid pixels are discarded. Remaining patches are grouped by geographic footprint and timestamp. If observations from multiple HSI sensors intersect the same footprint, they are grouped as co-located HSI views. This grouping produces locations with a single HSI sensor, paired HSI sensors, or all three HSI sensors, depending on acquisition availability.

The initial HSI archive is strongly imbalanced regarding sensor combinations. Before rebalancing, it contains approximately 4.1M \EMIT locations, 1.8M \EnMAP locations, and 275K \DESIS locations, corresponding to roughly 12M, 2.6M, and 447K HSI patches, respectively. The archive is dominated by \EMIT-only and \EMIT--\EnMAP pairs, while \DESIS coverage is comparatively limited.

\paragraph{Balancing HSI locations}
We rebalance the HSI archive before pairing it with the remaining EO modalities. Rebalancing is performed separately for different HSI co-location configurations, rather than applying a single global subsampling rule. We retain \DESIS-containing configurations because of their limited availability, and retain \EnMAP-only locations because they cover regions outside the ISS orbit. We downsample the two largest configurations: \EMIT-only locations are reduced from approximately 2.9M to 500K, and \EMIT--\EnMAP locations are reduced from approximately 1.1M to 750K.

We employ AlphaEarth embeddings \citep{brown2025alphaearth} for subsampling. For each HSI location, we select a reference year, using the majority year across available HSI observations when multiple sensors or timestamps are present. We then retrieve the corresponding annual AlphaEarth embeddings and average them to obtain one 64-dimensional embedding per location. These embeddings are clustered using a three-level hierarchical clustering procedure, and samples are selected within the resulting hierarchy following~\citep{voautomatic}. This reduces redundancy in overrepresented regions while preserving broad geographic and semantic coverage. All timestamps associated with selected locations are retained.

After rebalancing, the selected HSI anchor patches comprise approximately 1.4M \EnMAP locations, 1.4M \EMIT locations, and 275K \DESIS locations. These counts are not additive because a geographic location can contain observations from multiple HSI sensors and dates. \Cref{fig:dataset_global_coverage} shows the geographic distribution of these HSI anchor patches.

\begin{figure}[t]
  \centering
  \includegraphics[width=\linewidth]{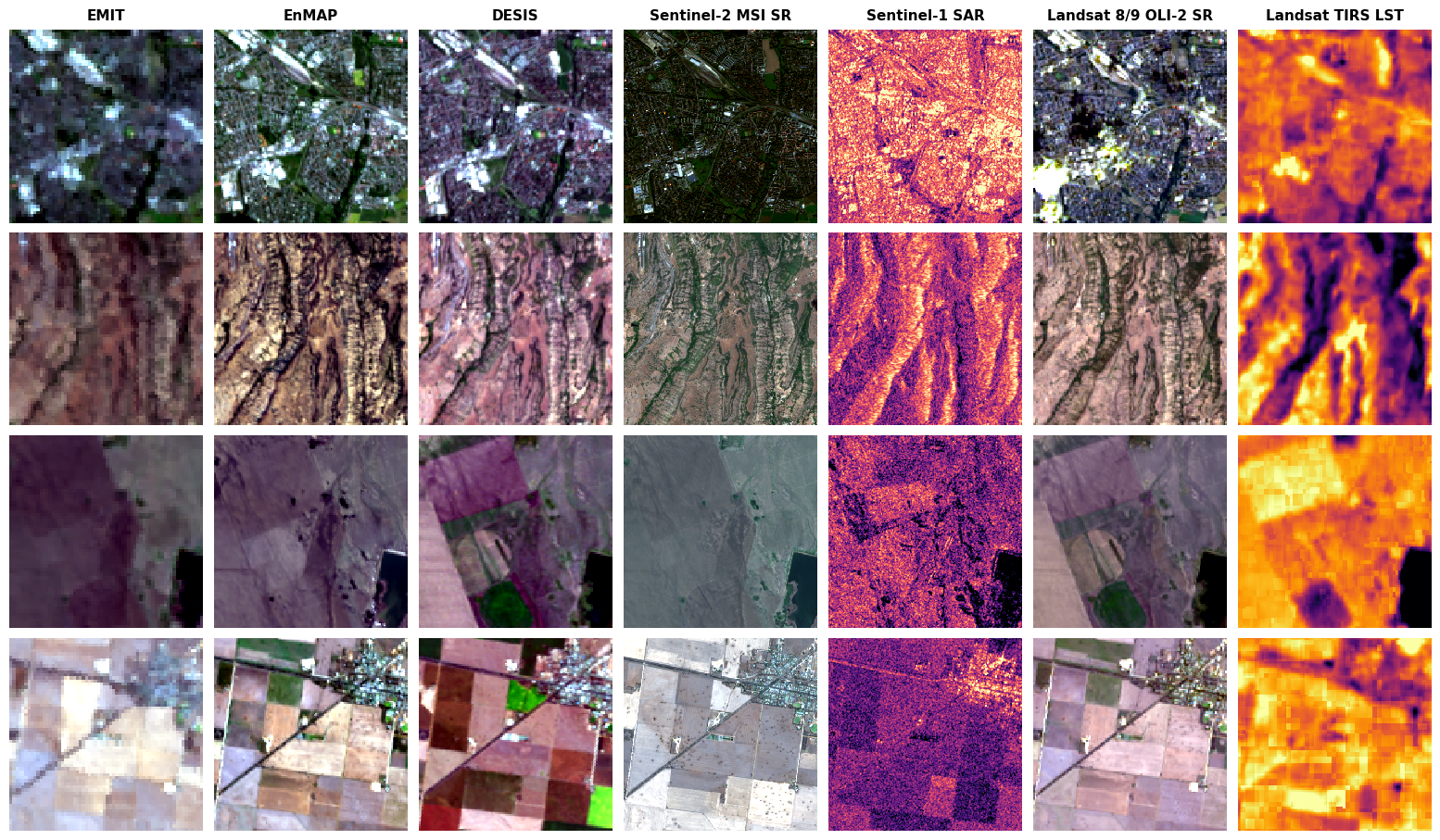}
  \caption{
  Examples of co-located observations in \datasetname. Each row corresponds to a geographic footprint and each column to an observation type. 
  }
  \label{fig:dataset_examples}
\end{figure}

\paragraph{Multisensor pairing}
After HSI rebalancing, each selected HSI footprint is paired with \Stwo, \Landsat optical imagery, Landsat LST, and \Sone observations. HSI timestamps are used as temporal anchors. For \Stwo and \Landsat optical imagery, we select observations within a maximum window of 6 months around the HSI anchor and apply a 5\% cloud threshold. For \Sone, we use a wider pairing window of up to 4 years to account for radar acquisition gaps. We retain only locations for which paired observations are available for the additional sensors along with at least one HSI anchor. Depending on HSI sensor availability, a location can therefore contain observations from five to seven sensor/product types. \Cref{fig:dataset_examples} illustrates four examples (rows) where all \datasetname sensors (columns) have been spatially co-located.

\paragraph{Temporal filtering and de-duplication}
Temporal coverage is uneven across locations and sensors. Some locations have dense time series, while others have only one or a few acquisitions. To reduce redundancy and bound storage and training cost, we retain at most four timestamps per sensor and location. Selection favors temporal proximity between paired cross-sensor observations and temporal spread across the year when available. We also remove near-duplicate observations from the same sensor acquired within 10 days of each other. The final dataset contains approximately 2M globally distributed locations, 25M georeferenced patches, and over 40 TB of data.

\subsection{Hyperspectral downstream tasks}

\paragraph{Co-location of HSI tasks with MSI}
To evaluate \modelname utilizing all optical branches, we extend the SpectralEarth \citep{braham2024spectralearth} downstream tasks \textit{CDL} (U.S. agriculture), \textit{BDForet} (French forests), and \textit{EuroCrops} (EU agriculture) with \Stwo and \Landsat imagery. In sync with the filtering of the HSI modalities in SpectralEarth, we filter the MSI by cloud coverage below 10 \% and a maximum temporal lag to the HSI modalities of 10 days for CDL and EuroCrops, and 45 days for BDForet.

\subsection{Spatio-spectral characteristics and licensing}
\label{sec:datalicenses}

\Cref{tab:licensesspatspec} summarizes spatio-spectral characteristics and corresponding data licenses for legal purposes.

\begin{table}[h!]
\centering
\resizebox{\columnwidth}{!}{%
\begin{tabular}{lp{1cm}cp{2.25cm}p{3.8cm}p{5.2cm}}
\hline
Mission & Operator & Spat.\ Res.\ [m] & Spectrum [$\mu$m] & Revisit [d] (note) & License \\ 
\hline
EMIT & NASA & 60 & 0.38--2.5 &\phantom{0}$\sim$4\newline(effective, ISS orbit and tasking constrained) & NASA Earth Science Data and Information Policy: fully open access without use restrictions \cite{nasa_esd_dis_policy} \\
DESIS & DLR & 30 & 0.4--1.0 &\phantom{0}$\sim$4\newline(effective, ISS orbit and tasking constrained) & DLR/Teledyne scientific-use license: free for non-commercial scientific research only \cite{desis_license} \\
EnMAP & DLR & 30 & 0.42--2.45 &\phantom{0}27\quad in nadir mode\newline \phantom{00}4\quad in off-nadir mode\newline(tasked acquisitions) & Free and open access via DLR geoservice upon registration; scientific and commercial use permitted \cite{enmap_geoservice} \\
Sentinel-2 & ESA & 10\slash20\slash60 & 0.44--2.19 & \phantom{00}5\newline(constellation; 10-day/unit) & Copernicus free, full and open access data policy \cite{copernicus_data_policy} \\
Landsat 8/9 & NASA\slash USGS & 30 & 0.43--2.29 (OLI)\newline10.6--12.5 (TIR) &\phantom{00}8\newline(constellation; 16-day/unit) & USGS Landsat Collection open data policy: unrestricted public access and redistribution \cite{usgs_landsat_data_policy} \\
Sentinel-1 & ESA & 10--20 & 0.056\newline(C-band SAR) & \phantom{00}6\newline(constellation; 12-day/unit) & Copernicus free, full and open access data policy \cite{copernicus_data_policy} \\
\hline
\end{tabular}
}
\caption{\label{tab:licensesspatspec}Summary of \datasetname's data sources and licensing conditions. Resolution denotes spatial ground sampling distance. Spectral ranges are given as wavelength intervals; SAR denotes synthetic aperture radar, TIR thermal infrared. ISS-based instruments (EMIT and DESIS) have no fixed orbital revisit; reported values correspond to effective revisit determined by orbital geometry and acquisition tasking constraints. Revisit times for satellite constellations refer to average global repeat cycles and may vary with latitude and observation mode.}
\end{table}

\section{Model details}
\label{app:model_details}

\paragraph{Spectral grouping and tokenization}
For HSI modalities, the retained valid bands of each sensor are partitioned into contiguous, evenly separated groups across the entire spectrum. The grouping is a hyperparameter and sensor-specific. Groups are defined over the retained band inventory, excluding gaps associated with atmospheric absorption windows. \citep{Bohnetal2024}. EnMAP \citep{DLR_EnMAP_FAQ} uses 15 groups, DESIS \citep{Alonsoetal2019} uses 10 groups, and EMIT \citep{NASA_EMIT_L2A_Guide} uses 15 groups. The average group
sizes are 13.5, 21.5, and 16.3 bands, respectively, with group sizes ranging
from 8 to 23 bands across the HSI sensors.

Each spectral group is embedded independently into a spectral token dimension
$d_\lambda=96$. A four-layer spectral transformer with two attention heads
processes the group-token sequence at each spatial position. Group tokens are
then aggregated with projected-attention pooling: they are projected to a
spectral fusion dimension of 192, attended by a learned query with two heads,
and mapped to the common spatial embedding dimension $D=192$.

\paragraph{Patch embedding and token layout}
Sensor-specific patch projections are chosen so that inputs for all modalities produce a
common $64\times64$ spatial token grid before the local Hiera branches. In
native sensor pixels, the effective patch strides are $2\times2$ for EnMAP,
DESIS, Landsat optical (OLI), and Landsat LST; $1\times1$ for EMIT; and $6\times6$
for Sentinel-2 and Sentinel-1. The corresponding input crops are
$128\times128$ for EnMAP, DESIS, Landsat optical, and Landsat LST,
$64\times64$ for EMIT, and $384\times384$ for Sentinel-2 and Sentinel-1.
The convolutional patch projections use overlapping kernels for most
modalities: $3\times3$ for the $2\times2$ streams, $1\times1$ for EMIT, and
$7\times7$ for the $10~m$ sensors.

\paragraph{Hiera configuration}
The multimodal backbone uses a three-stage Hiera~\citep{ryali2023hiera} layout with depths
$[2,6,10]$. The initial spatial token embedding size is $D=192$, with two attention
heads. Width and head count are doubled at each pooling transition. The first
two Hiera stages are sensor-specific local branches. Stage 1 keeps $D=192$
and operates on the $64\times64$ token grid. Stage 2 begins with a
$2\times2$ query-pooling transition, reducing the token grid to
$32\times32$ and increasing the width to $D_\ell=384$ with four heads. The local branch have shape
${32^2 \times 384}$ before fusion.

The shared Hiera trunk corresponds to the final stage. Its first block applies
the second $2\times2$ query-pooling transition, reducing the grid to
$16\times16$ and increasing the width to 768 with eight heads. The remaining
shared blocks operate at this final resolution and width. The reported
configuration does not use a class token; image-level representations are
obtained by mean-pooling the final spatial tokens. Dense prediction heads can
instead consume intermediate hierarchical feature maps.

\paragraph{Fusion configuration}
Cross-sensor fusion is applied after the local branches and before the shared
Hiera stage. For each spatial token position, available sensor embeddings
$H_s \in \mathbb{R}^{D_\ell}$ are projected from $D_\ell=384$ to a fusion
dimension $D_f=768$. Learned sensor embeddings are added in this fusion space,
and a learned query aggregates the available sensor tokens using two-head
projected attention. The fused output is projected back to $D_\ell=384$,
yielding a fused matrix of shape ${32^2 \times 384}$, which is the input expected by the shared Hiera trunk. The same projected-attention fusion path is used for multimodal and single-stream views.

\paragraph{Model size}
The multimodal Hiera encoder contains approximately 156M trainable parameters.
Most parameters are in the sensor-specific local Hiera branches and the shared
Hiera trunk, while the spectral tokenizers and fusion module are comparatively
small. 

\section{Training details}
\label{app:training_details}

\paragraph{Optimization}
Pretraining uses AdamW with weight decay $0.05$, a peak learning rate of
$10^{-5}$, and an effective batch size of $240$ distributed over $48$ A100
GPUs. The learning rate is linearly warmed up and then decayed with a cosine
schedule. The EMA teacher momentum follows a cosine schedule from $0.996$ to
$1.0$ over training. We use mixed-precision training. The LeJEPA \citep{balestriero2025lejepa} projection 
head maps image representations to dimension
$128$ with a three-layer MLP, and the SIGReg term uses $\lambda=0.015$.

\paragraph{View construction}
For each location, pretraining constructs global and local augmented views from
the available co-located sensor streams. Each global view contains four sensors sampled 
independently from the available modalities, and global crops
are sampled with scale range $[0.4, 1.0]$. Local views contain a single sensor,
sampled from the modalities present in the global views, with crop scale
$[0.1, 0.4]$. Spatial
augmentations are applied consistently across the streams within a view, while
radiometric jitter and Gaussian blur are applied independently to each sensor.

\paragraph{Downstream training}
For hyperspectral downstream tasks, we use the lightweight decoder described 
in \Cref{sec:hsi_eval}: selected encoder feature maps are projected to dimension $128$, 
upsampled to the largest
feature resolution, concatenated, refined with a $3\times3$ convolution, and
mapped to class logits with a $1\times1$ classifier. Unless otherwise stated,
the head is trained for $100$ epochs with AdamW, cosine learning-rate decay,
weight decay $10^{-4}$, and learning rate $10^{-4}$.

\end{document}